\documentclass[twoside,11pt]{article}
\usepackage[bookmarks=true]{hyperref}
\hypersetup{%
    pdftitle={Autotelic Agents with Intrinsically Motivated Goal-Conditioned RL: A Short Survey},    % title
    pdfauthor={Cédric Colas, Tristan Karch, Olivier Sigaud, Pierre-Yves Oudeyer},                     % author
    pdfsubject={Shorty survey on Autotelic RL},  % subject of the document
    pdfkeywords={autotelic, intrinsic motivations, reinforcement learning, artificial intelligence}, % list of keywords
    colorlinks=true,       % false: boxed links; true: colored links
    linkcolor=black,       % color of internal links
    citecolor=black,       % color of links to bibliography
    filecolor=black,        % color of file links
    urlcolor=black,        % color of external links
    linktoc=page            % only page is linked
}
\usepackage{jair, theapa, rawfonts}
\jairheading{74}{2022}{1159-1199}{12/2021}{07/2022}
\ShortHeadings{autotelic agents with intrinsically motivated goal-conditioned rl}
{Colas, Karch, Sigaud \& Oudeyer}
\firstpageno{1159}

\usepackage[dvipsnames]{xcolor}
\usepackage{float}
\definecolor{myred}{rgb}{0.8,0,0}
\definecolor{mygreen}{rgb}{0,0.6,0}
\definecolor{myblue}{rgb}{0,0,0.7}
\usepackage{tcolorbox}
\usepackage{lmodern}% http://ctan.org/pkg/lmodern
\usepackage{slantsc}% http://ctan.org/pkg/slantsc

\newcommand{\todo}[1]{}
\renewcommand{\todo}[1]{{\color{myred} Todo: {#1}}}
\newcommand{\m}[1]{\mathcal{#1}}
\usepackage{amsfonts}
\usepackage{amsmath}
\usepackage{dsfont}
\usepackage[noend]{algpseudocode} 
\usepackage{algorithm}
\usepackage{multirow}
\usepackage{graphicx}
\usepackage{lscape}
\usepackage{xspace}
\usepackage{array}
\usepackage{varwidth}

\newcommand{\ie}{i.e.\,}
\newcommand{\eg}{e.g.\,}

\newcommand{\rl}{\textsc{rl}\xspace}
\newcommand{\ai}{\textsc{ai}\xspace}
\newcommand{\drl}{\textsc{drl}\xspace}
\newcommand{\hrl}{\textsc{hrl}\xspace}

\newcommand{\popimgep}{\textsc{pop-imgep}\xspace}
\newcommand{\rlimgep}{\textsc{rl-imgep}\xspace}
\newcommand{\rlimgeps}{\textsc{rl-imgep}s\xspace}
\newcommand{\rlemgep}{\textsc{rl-emgep}\xspace}
\newcommand{\rlemgeps}{\textsc{rl-emgep}s\xspace}
\newcommand{\imgep}{\textsc{imgep}\xspace}
\newcommand{\imgeps}{\textsc{imgep}s\xspace}
\newcommand{\uvfa}{\textsc{uvfa}\xspace}
\newcommand{\uvfas}{\textsc{uvfa}s\xspace}
\newcommand{\horde}{\textsc{horde}\xspace}
\newcommand{\imagine}{\textsc{imagine}\xspace}

\newcommand{\rnn}{\textsc{rnn}\xspace}

\newcommand{\gan}{\textsc{gan}\xspace}
\newcommand{\vae}{\textsc{vae}\xspace}
\newcommand{\lp}{\textsc{lp}\xspace}

\newcommand{\mdp}{\textsc{mdp}\xspace}
\newcommand{\mdps}{\textsc{mdp}s\xspace}
\newcommand{\ims}{\textsc{im}s\xspace}
\newcommand{\im}{\textsc{im}s\xspace}

\makeatletter
\newcommand{\algmargin}{\the\ALG@thistlm}
\makeatother

\newlength{\whilewidth}
\algdef{SE}[parWHILE]{parWhile}{EndparWhile}[1]
  {\parbox[t]{\dimexpr\linewidth-\algmargin}{%
     \hangindent\whilewidth\strut\algorithmicwhile\ #1\ \algorithmicdo\strut}}{\algorithmicend\ \algorithmicwhile}%
\algnewcommand{\parState}[1]{\State%
  \parbox[t]{\dimexpr\linewidth-\algmargin}{\strut #1\strut}}
  
\makeatletter
\newlength{\trianglerightwidth}
\algnewcommand{\LineComment}[1]{\Statex \hskip\ALG@thistlm $\triangleright$ #1}
\algnewcommand{\LineCommentCont}[1]{\Statex \hskip\ALG@thistlm%
  \parbox[t]{\dimexpr\linewidth-\ALG@thistlm}{\hangindent=\trianglerightwidth \hangafter=1 \strut$~~~\triangleright$ #1\strut}}
\algnewcommand{\LineCommentConttwo}[1]{\Statex \hskip\ALG@thistlm%
  \parbox[t]{\dimexpr\linewidth-\ALG@thistlm}{\hangindent=\trianglerightwidth \hangafter=1 \strut$~~~~~\nobreak\hspace{.16667em plus .08333em}~~\triangleright$ #1\strut}}
\makeatother

\begin{document}

\title{Autotelic Agents with Intrinsically Motivated Goal-Conditioned Reinforcement Learning:\\ A Short Survey}

%\author{}
\author{\name C\'edric Colas \email cedric.colas@inria.fr \\
        \addr INRIA and Univ. de Bordeaux; Bordeaux (FR)\\
        \AND
        \name Tristan Karch \email tristan.karch@inria.fr \\
        \addr INRIA and Univ. de Bordeaux; Bordeaux (FR)\\
        \AND
        \name Olivier Sigaud \email olivier.sigaud@upmc.fr \\
        \addr Sorbonne Université; Paris (FR) \\
        \AND
        \name Pierre-Yves Oudeyer \email pierre-yves.oudeyer@inria.fr \\
        \addr INRIA; Bordeaux (FR) and ENSTA Paris Tech; Paris (FR)}

% For research notes, remove the comment character in the line below.
% \researchnote

\maketitle

\begin{abstract}
    Building autonomous machines that can explore open-ended environments, discover possible interactions and build repertoires of skills is a general objective of artificial intelligence. Developmental approaches argue that this can only be achieved by \textit{autotelic agents}: intrinsically motivated learning agents that can learn to represent, generate, select and solve their own problems. 
    In recent years, the convergence of developmental approaches with deep reinforcement learning (\rl) methods has been leading to the emergence of a new field: \textit{developmental reinforcement learning}. 
    Developmental \rl is concerned with the use of deep \rl algorithms to tackle a developmental problem\,---\,the \textit{intrinsically motivated acquisition of open-ended repertoires of skills}.
    The self-generation of goals requires the learning of compact goal encodings as well as their associated goal-achievement functions. This raises new challenges compared to standard \rl algorithms originally designed to tackle pre-defined sets of goals using external reward signals.
    The present paper introduces developmental \rl and proposes a computational framework based on goal-conditioned \rl to tackle the intrinsically motivated skills acquisition problem. It proceeds to present a typology of the various goal representations used in the literature, before reviewing existing methods to learn to represent and prioritize goals in autonomous systems. We finally close the paper by discussing some open challenges in the quest of intrinsically motivated skills acquisition.
\end{abstract}

% % % % % % % % % % % % % % % % % % % % % % % % % % %
% INTRO
% % % % % % % % % % % % % % % % % % % % % % % % % % % 
\section{Introduction}
\label{Introduction}

% Human learning is great, we’d like to train agents to show similar learning abilities and behavior
Building autonomous machines that can explore large environments, discover interesting interactions and learn open-ended repertoires of skills is a long-standing goal in artificial intelligence. Humans are remarkable examples of this lifelong, open-ended learning. They learn to recognize objects and crawl as infants, then learn to ask questions and interact with peers as children. Across their lives, humans build a large repertoire of diverse skills from a virtually infinite set of possibilities. What is most striking, perhaps, is their ability to invent and pursue their own problems, using internal feedback to assess completion. We would like to build artificial agents able to demonstrate equivalent lifelong learning abilities.

% We can think of two fields targeting the problem of learning behaviors

We can think of two approaches to this problem: developmental approaches, in particular developmental robotics, and reinforcement learning (\rl). Developmental robotics takes inspirations from artificial intelligence, developmental psychology and neuroscience to model cognitive processes in natural and artificial systems \shortcite{asada2009cognitive,cangelosi2015developmental}. Following the idea that intelligence should be \textit{embodied}, robots are often used to test learning models. Reinforcement learning, on the other hand, is the field interested in problems where agents learn to behave by experiencing the consequences of their actions under the form of rewards and costs. As a result, these agents are not explicitly taught, they need to learn to maximize cumulative rewards over time by trial-and-error \shortcite{sutton2018reinforcement}. While developmental robotics is a field oriented towards answering particular questions around sensorimotor, cognitive and social development (\eg how can we model language acquisition?), reinforcement learning is a field organized around a particular technical framework and set of methods.

% RL - Standard, multi-goal, limits, awesome achievements

Now powered by deep learning optimization methods leveraging the computational efficiency of large computational clusters, \rl algorithms have recently achieved remarkable results including, but not limited to, learning to solve video games at a super-human level \shortcite{mnih2015human}, to beat chess and go world players \shortcite{silver2016mastering}, or even to control stratospheric balloons in the real world \shortcite{bellemare2020autonomous}.

Although standard \rl problems often involve a single agent learning to solve a unique task, \rl researchers  extended \rl problems to \textit{multi-goal \rl problems}. Instead of pursuing a single goal, agents can now be trained to pursue goal distributions \shortcite{kaelbling1993learning,sutton2011horde,schaul2015universal}. As the field progresses, new goal representations emerge: from the specific goal states to the high-dimensional goal images or the abstract language-based goals \shortcite{Luketina2019}. However, most approaches still fall short of modeling the learning abilities of natural agents because they train them to solve predefined sets of tasks, via external and hand-defined learning signals.

% Dev robotics - good approach, formulate the right problem, but not awesome achievements
Developmental robotics directly aims to model children learning and, thus, takes inspiration from the mechanisms underlying autonomous behaviors in humans. Most of the time, humans are not motivated by external rewards but spontaneously explore their environment to discover and learn about what is around them. This behavior seems to be driven by \textit{intrinsic motivations} (\ims) a set of brain processes that motivate humans to explore for the mere purpose of experiencing novelty, surprise or learning progress \shortcite{berlyne1966curiosity,gopnik1999scientist,kidd2015psychology,oudeyer2016evolution,gottlieb2018towards}.

The integration of \ims into artificial agents thus seems to be a key step towards autonomous learning agents \shortcite{schmidhuber1991possibility,kaplan2007search}. In developmental robotics, this approach enabled sample efficient learning of high-dimensional motor skills in complex robotic systems \shortcite{santucci2020intrinsically}, including locomotion \shortcite{baranes2013active,martius2013information}, soft object manipulation \shortcite{rolf2013efficient,nguyen2014socially}, visual skills \shortcite{lonini2013robust} and nested tool use in real-world robots \shortcite{imgep}. Most of these approaches rely on \textit{population-based} optimization algorithms, non-parametric models trained on datasets of (policy, outcome) pairs. Population-based algorithms cannot leverage automatic differentiation on large computational clusters, often demonstrate limited generalization capabilities and cannot easily handle high-dimension perceptual spaces (\eg images) without hand-defined input pre-processing. For these reasons, developmental robotics could benefit from new advances in deep \rl.

% Recently, we’ve been observing a convergence between the two with ideas from dev rob being integrated within RL
%     frameworks, and RL tools being used for dev rob objectives.
Recently, we have been observing a convergence of these two fields, forming a new domain that we propose to call \textit{developmental reinforcement learning}, or more broadly \textit{developmental artificial intelligence}. Indeed, \rl researchers now incorporate fundamental ideas from the developmental robotics literature in their own algorithms, and reversely developmental robotics learning architecture are beginning to benefit from the generalization capabilities of deep \rl techniques. These convergences can mostly be categorized in two ways depending on the type of intrinsic motivation (\ims) being used \shortcite{oudeyer2007intrinsic}:
\begin{itemize}
    \item
    \textbf{Knowledge-based IMs} are about prediction. They compare the situations experienced by the agent to its current knowledge and expectations, and reward it for experiencing dissonance (or resonance). This family includes \ims rewarding prediction errors \shortcite{schmidhuber1991possibility,pathak2017curiosity}, novelty \shortcite{bellemare2016unifying,burda2018exploration,raileanu2020ride}, surprise \shortcite{achiam2017surprise}, negative surprise \shortcite{berseth2019smirl}, learning progress \shortcite{lopes2012exploration,kim2020active} or information gains \shortcite{houthooft2016vime}, see a review in \shortciteA{linke2019adapting}. This type of \im is often used as an auxiliary reward to organize the exploration of agents in environments characterized by sparse rewards. It can also be used to facilitate the construction of world models \shortcite{lopes2012exploration,kim2020active,sekar2020planning}.
    \item
    \textbf{Competence-based IMs}, on the other hand, are about control. They reward agents to solve self-generated problems, to achieve self-generated goals. In this category, agents need to represent, select and master self-generated goals. As a result, competence-based \ims were often used to organize the acquisition of repertoires of skills in task-agnostic environments \shortcite{baranes2010intrinsically,baranes2013active,santucci2016grail,forestier2016modular,nair2018visual,warde2018unsupervised,curious,blaes2019control,pong2019skew,imagine}. Just like knowledge-based \ims, competence-based \ims organize the exploration of the world and, thus, might be used to train world models \shortcite{baranes2013active,chitnis2020glib} or facilitate learning in sparse reward settings \shortcite{geppg}. We propose to use the adjective \textbf{autotelic}, from the Greek \textit{auto} (self) and \textit{telos} (end, goal), to characterize agents that are intrinsically motivated to represent, generate, pursue and master their own goals (\ie that are both intrinsically motivated and goal-conditioned).
\end{itemize}

\rl algorithms using \textit{knowledge-based} \ims leverage ideas from developmental robotics to solve standard \rl problems. On the other hand, \rl algorithms using competence-based \ims organize exploration around self-generated goals and can be seen as targeting a developmental robotics problem: the \textit{open-ended and self-supervised acquisition of repertoires of diverse skills}. 
 
\textit{Intrinsically Motivated Goal Exploration Processes} (\imgep) is the family of autotelic algorithms that bake competence-based \ims into learning agents \shortcite{imgep}. \imgep agents generate and pursue their own goals as a way to explore their environment, discover possible interactions and build repertoires of skills. This framework emerged from the field of developmental robotics \shortcite{oudeyer2007intrinsic,baranes2009proximo,baranes2010intrinsically,rolf2010goal} and originally leveraged population-based learning algorithms (\popimgep) \shortcite{baranes2009r,baranes2013active,forestier2016modular,imgep}. 

Recently, goal-conditioned \rl agents were also endowed with the ability to generate and pursue their own goals and learn to achieve them via self-generated rewards. We call this new set of autotelic methods \rlimgeps. In contrast, one can refer to externally-motivated goal-conditioned \rl agents as \rlemgeps.

\begin{figure}[h]
    \centering
    \includegraphics[width=\textwidth]{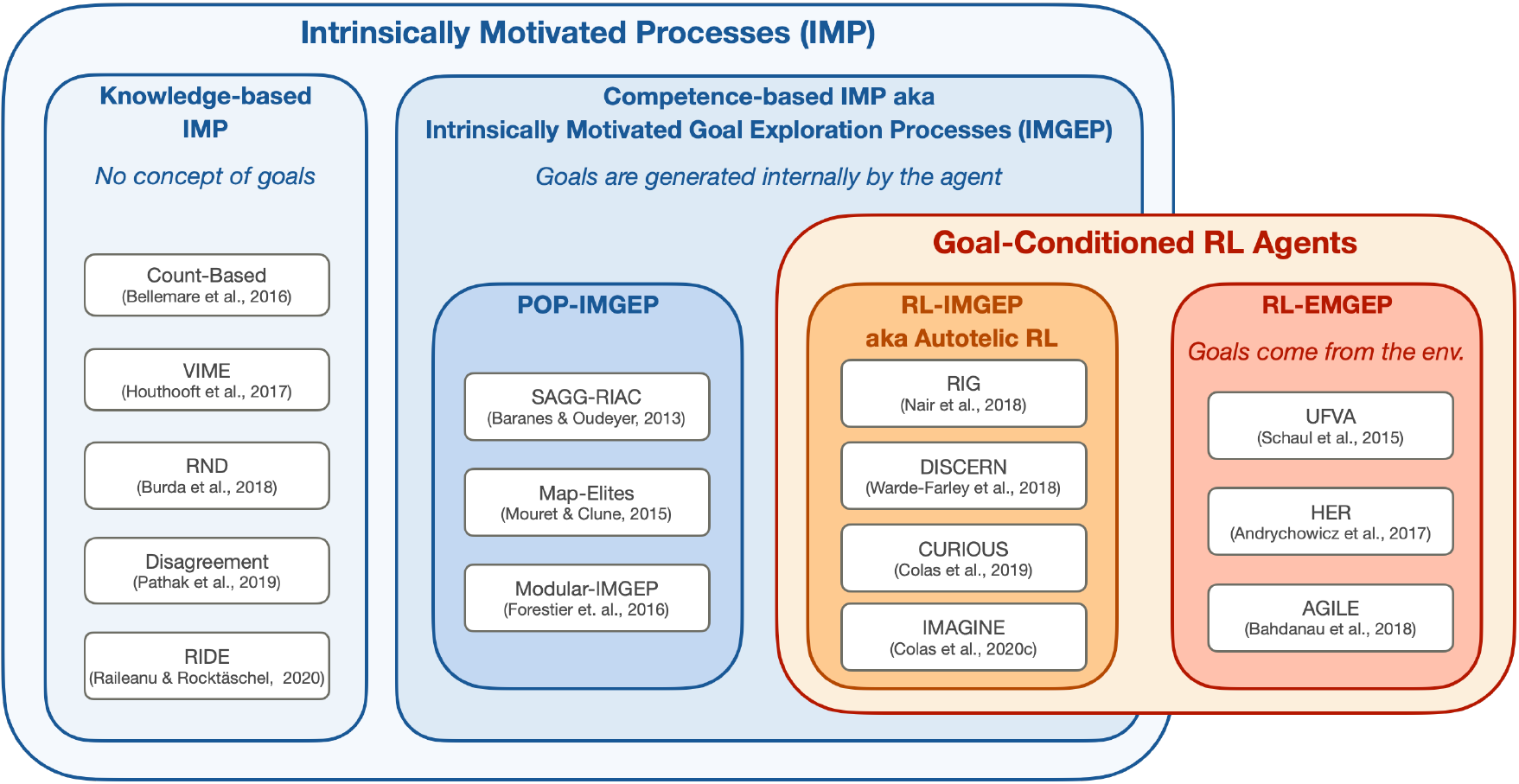}
    \caption{A typology of intrinsically-motivated and/or goal-conditioned \rl approaches. \popimgep, \rlimgep
    and \rlemgep refer to \textit{population-based intrinsically motivated goal exploration processes}, 
        \textit{\textsl{\rl}-based} \textsl{\imgep} and \textit{\textsl{\rl}-based externally motivated goal exploration
        processes} respectively. \popimgep, \rlimgep and \rlemgep all represent goals, but knowledge-based \ims
        do not. While \imgeps (\popimgep and \rlimgep) generate their own goals, \rlemgeps require
        externally-defined goals. This paper is interested in \rlimgeps, autotelic methods at the intersection of
        \textit{goal-conditioned \rl agents} and \textit{intrinsically motivated processes} that train learning
        agents to generate and pursue their own goals with goal-conditioned \rl algorithms. }
    \label{fig:scope}
\end{figure}

% This is what this survey is about
This paper proposes a formalization and a review of the \rlimgep algorithms at the convergence of \rl methods and developmental robotics objectives. Figure~\ref{fig:scope} proposes a visual representation of intrinsic motivations approaches (knowledge-based \ims vs competence-based \ims or \imgeps) and goal-conditioned \rl (externally vs intrinsically motivated). Their intersection is the family of autotelic algorithms that train agents to generate and pursue their own goals by training goal-conditioned policies. 

We define goals as the combination of a compact goal representation and a goal-achievement function to measure progress. This definition highlights new challenges for autonomous learning agents. While traditional \rl agents only need to learn to achieve goals, \rlimgep agents also need to learn to represent them, to generate them and to measure their own
progress. After learning, the resulting goal-conditioned policy and its associated goal space form a \textit{repertoire of skills}, a repertoire of behaviors that the agent can represent and control. We believe organizing past goal-conditioned \rl algorithms at the convergence of developmental robotics and \rl into a common classification and towards the resolution of a common problem will help organize future research.

\begin{tcolorbox}
\small
\paragraph{Definitions}
\begin{itemize}
    \item \textit{\textbf{Goal}}: ``a cognitive representation of a future object that the organism is committed to approach \shortcite{elliot2008goal}.'' In \rl, this takes the form of a (embedding, goal-achievement function) pair, see Section~\ref{sec:goals}.
    \item \textit{\textbf{Skill}}: the association of a goal and a policy to reach it, see Section~\ref{sec:im_pb}.
    \item \textit{\textbf{Goal-achievement function}}: a function that measures progress towards a goal (also called goal-conditioned reward function), see Section~\ref{sec:goals}.
    \item \textit{\textbf{Goal-conditioned policy}}: a function that generates the next action given the current state and the goal, see Section~\ref{sec:im_pb_solution}.
    \item \textit{\textbf{Autotelic}}: from the Greek \textit{auto} (self) and \textit{telos} (end, goal), characterizes agents that generate their own goals and learning signals. In is equivalent to \textit{intrinsically motivated and goal-conditioned}.
\end{itemize}
\end{tcolorbox}

\paragraph{Scope of the survey.}
We are interested in algorithms from the \rlimgep family as algorithmic tools to enable agents to acquire repertoires of skills in an open-ended and self-supervised setting. Externally motivated goal-conditioned \rl approaches do not enable agents to generate their own goals and thus cannot be considered autotelic (\imgeps). However, these approaches can often be converted into autotelic \rlimgeps by integrating the goal generation process within the agent. For this reason, we include some \rlemgeps approaches when they present interesting mechanisms that can directly be leveraged in autotelic agents. 

%what’s not covered and why
\paragraph{What is not covered.} This survey does not discuss some related but distinct approaches such as multi-task \rl \shortcite{caruana1997multitask}, \rl with auxiliary tasks \shortcite{riedmiller2018learning,jaderberg2016reinforcement} and \rl with knowledge-based \ims \shortcite{bellemare2016unifying,pathak2017curiosity,burda2018exploration}. None of these approaches do represent goals or see the agent's behavior affected by goals. The subject of intrinsically motivated goal-conditioned \rl also relates to \textit{transfer learning} and \textit{curriculum learning}. This survey does not cover transfer learning approaches, but interested readers can refer to \shortciteA{taylor2009transfer}. It  discusses automatic curriculum learning approaches that organize the generation of goals according to the agent's abilities in Section~\ref{sec:survey_generation} but, for a broader picture on the topic, readers can refer to the recent review \shortciteA{portelas2020automatic}. Finally, this survey does not review policy learning methods but only focuses on goal-related mechanisms. Indeed, the choice of mechanisms to learn to represent and select goals is somewhat orthogonal to the algorithms used to learn to achieve them. Since the policy learning algorithms used in \rlimgep architectures do not differ significantly from standard \rl and goal-conditioned \rl approaches, this survey focuses on goal-related mechanisms, specific to \rlimgeps.

\paragraph{Survey organization.} We start by presenting some background on the formalization of \rl and multi-goal \rl problems and the corresponding algorithms to solve them (Section~\ref{sec:background}). We then build on these foundations to formalize the \textit{intrinsially motivated skills acquisition problem} and propose a computational framework to tackle it: \textit{\textsl{\rl}-based intrinsically motivated goal exploration processes} (Section~\ref{sec:im_pb_solution}). Once this is done, we organize the surveyed literature along three axes: 1) What are the different types of goal representations? (Section~\ref{sec:survey_goal_rep}); 2) How can we learn goal representations? (Section~\ref{sec:survey_learning_goal_rep}) and 3) How can we prioritize goal selection? (Section~\ref{sec:survey_generation}). We finally close the survey on a discussion of open challenges for developmental reinforcement learning (Section~\ref{sec:future}).

%\os{On pourrait peut-être mentionner qu'on ne couvre pas le cas (rarement étudié) où chaque tâche utilise sa propre
%    state action representation (cf. \shortcite{doncieux2018open}).}

% % % % % % % % % % % % % % % % % % % % % % % % % % %
% The Intrinsically Motivated Goal Exploration Problem
% % % % % % % % % % % % % % % % % % % % % % % % % % % 

\section{Background: RL, Multi-Goal RL Problems and Their Solutions}
\label{sec:background}
This sections presents background information on the \rl problem, the multi-goal \rl problem and the families of algorithms used to solve them. This will serve as a foundation to define the \textit{intrinsically motivated skill acquisition problem} and introduce the  \textit{\textsl{\rl}-based intrinsically motivated goal exploration process} framework to solve it (\rlimgep, Section~\ref{sec:im_pb_solution}).

\subsection{The Reinforcement Learning Problem}
In a reinforcement learning (\rl) problem, the agent learns to perform sequences of actions in an environment so as to maximize some notion of cumulative reward \shortcite{sutton2018reinforcement}. \rl problems are commonly framed as Markov Decision Processes (\mdps): $\m{M}\,=\,\{\m{S},\,\m{A},\,\m{T},\,\rho_0,\,R\}$ \shortcite{sutton2018reinforcement}. The agent and its environment, as well as their interaction dynamics are defined by the first components $\{\m{S},\,\m{A},\,\m{T},\,\rho_0\}$, where $s\in\m{S}$ describes the current state of the agent-environment interaction and $\rho_0$ is the distribution over initial states. The agent can interact with the environment through actions $a\in\m{A}$. Finally, the dynamics are characterized by the transition function $\m{T}$ that dictates the distribution of the next state $s'$ from the current state and action $\m{T} (s'\mid s,\,a)$. 

The objective of the agent in this environment is defined by the remaining component of the \mdp: $R$. $R$ is the reward function, it computes a reward for any transition: $R(s,\,a,\,s')$. Note that, in a traditional \rl problem, the agent only receives the rewards corresponding to the transitions it experiences but does not have access to the function itself. The objective of the agent is to maximize the cumulative reward computed over complete episodes. When computing the aggregation of rewards, we often introduce discounting and give smaller weights to delayed rewards. $R^\text{tot}_t$ is then computed as $R^\text{tot}_t\,=\,\sum_{i=t}^\infty \gamma^{i-t}  R(s_{i-1},\,a_i,\,s_i)$ with $\gamma$ being a constant discount factor in $]0,\,1]$. Each instance of an \mdp implements an \rl problem, also called a \textit{task}.

\subsection{Defining \textit{Goals} for Reinforcement Learning}
\label{sec:goals}
This section takes inspiration from the notion of \textit{goal} in psychological research to inform the formalization of \textit{goals} for reinforcement learning.

\paragraph{\textit{Goals} in psychological research.} Working on the origin of the notion \textit{goal} and its use in past psychological research, \shortciteA{elliot2008goal} propose a general definition:
\begin{quote}
    \textit{A goal is a cognitive representation of a future object that the organism is committed to approach or
    avoid} \shortcite{elliot2008goal}.
\end{quote}
Because goals are \textit{cognitive representations}, only animate organisms that represent goals qualify as goal-conditioned. Because this representation relates to a \textit{future object}, goals are cognitive imagination of future possibilities: goal-conditioned behavior is proactive, not reactive. Finally, organisms \textit{commit} to their goal, their behavior is thus influenced directly by this cognitive representation.

\paragraph{Generalized \textit{goals} for reinforcement learning.} \rl algorithms seem to be a good fit to train such goal-conditioned agents. Indeed, \rl algorithms train learning agents (\textit{organisms}) to maximize (\textit{approach}) a cumulative (\textit{future}) reward (\textit{object}). In \rl, goals can be seen as a set of \textit{constraints} on one or several consecutive states that the agent seeks to respect. These constraints can be very strict and characterize a single target point in the state space (\eg image-based goals) or a specific sub-space of the state space (\eg target x-y coordinate in a maze, target block positions in manipulation tasks). They can also be more general, when expressed by language for example (\eg '\textit{find a red object or a wooden one}'). 

To represent these goals, \rl agents must be able to 1)~have a compact representation of them and 2)~assess their progress towards it. This is why we propose the following formalization for \rl goals: each goal is a $g\,=\,(z_g,\,R_g)$ pair where $z_g$ is a compact \textit{goal parameterization} or \textit{goal embedding} and $R_g$ is a \textit{goal-achievement} function measuring progress towards the goal. The set of goal-achievement function can be represented as a single \textit{goal-parameterized} or \textit{goal-conditioned} reward function such that $R_\m{G}(\cdot\mid\,z_g)\,=\,R_g(\cdot)$. With this definition we can express a diversity of goals, see Section~\ref{sec:survey_goal_rep} and Table~\ref{tab:bigtable}.

The goal-achievement function and the goal-conditioned policy both assign \textit{meaning} to a goal. The former defines what it means to achieve the goal, it describes how the world looks like when it is achieved. The latter characterizes the process by which this goal can be achieved; what the agent needs to do to achieve it. In this search for the meaning of a goal, the goal embedding can be seen as the map: the agent follows this map and via the two functions above, experiences the meaning of the goal.

\begin{tcolorbox}
\small
\paragraph{Generalized definition of the goal construct for RL:}
\begin{itemize}
    \item \textbf{\textit{Goal}}: a $g\,=\,(z_g,\,R_g)$ pair where $z_g$ is a compact \textit{goal parameterization} or \textit{goal embedding} and $R_g$ is a \textit{goal-achievement} function.
    \item \textbf{Goal-achievement function: $R_g(\cdot)\,=\,R_\m{G}(\cdot\mid\,z_g)$} where $R_\m{G}$ is a goal-conditioned reward function.
\end{itemize}
\end{tcolorbox}

\subsection{The Multi-Goal Reinforcement Learning Problem}
By replacing the unique reward function $R$ by the space of reward functions $\m{R}_\m{G}$, \rl problems can be extended to handle multiple goals: $\m{M}\,=\,\{\m{S},\,\m{A},\,\m{T},\,\rho_0,\,\m{R}_\m{G}\}$. The term \textit{goal} should not be mistaken for the term \textit{task}, which refers to a particular \mdp instance. As a result, \textit{multi-task} \rl refers to \rl algorithms that tackle a set of \mdps that can differ by any of their components (\eg $\m{T}$,\,$R$,\,$\m{S}_0$, etc.). The \textit{multi-goal} \rl problem can thus be seen as the particular case of the multi-task \rl problem where \mdps differ by their reward functions. In the standard multi-goal \rl problem, the set of goals\,---\,and thus the set of reward functions\,---\,is pre-defined by engineers. The experimenter sets goals to the agent, and provides the associated reward functions. 

\subsection{Solving the RL Problem with RL Algorithms and Related Approaches}
The \rl problem can be tackled by several types of optimization methods. In this survey, we focus on \rl
algorithms, as they currently demonstrate stronger capacities in multi-goal problems \shortcite{goalgan,eysenbach2018diversity,warde2018unsupervised,pong2019skew,lynch2020grounding,hill_human_2020,hill_grounded_2020,abramson_imitating_2020,imagine,team2021open}.

\rl algorithms use transitions collected via interactions between the agent and its environment $(s,\,a,\,s',\,R(s,\,a,\,s'))$ to train a \textit{policy} $\pi$: a function generating the next action $a$ based on the current state $s$ so as to maximize a cumulative function of rewards. Deep \rl  (\drl) is the extension of \rl algorithms that leverage deep neural networks as function approximators to represent policies, reward and value functions. It has been powering most recent breakthrough in \rl  \shortcite{eysenbach2018diversity,warde2018unsupervised,goalgan,pong2019skew,lynch2020grounding,hill_human_2020,hill_grounded_2020,abramson_imitating_2020,imagine,team2021open}.

Other sets of methods can also be used to train policies. Imitation Learning (\textsc{il}) leverages demonstrations, \ie transitions collected by another entity \shortcite<\eg >{ho2016generative,hester2018deep}. Evolutionary Computing (\textsc{ec}) is a group of population-based approaches where populations of policies are trained to maximize cumulative rewards using episodic samples \shortcite<\eg >{sehnke2010parameter,lehman2011evolving,wierstra2014natural,mouret2015illuminating,salimans2017evolution,imgep,colas2020scaling}. Finally, in model-based \rl approaches, agents learn a model of the transition function $\m{T}$. Once learned, this model can be used to perform planning towards reward maximization or train a policy via \rl using imagined samples (\eg \shortciteB{schmidhuber_making_1990,dayan_helmholtz_1995,nguyen-tuong_model_2011,chua2018deep,charlesworth2020plangan,schrittwieser_mastering_2020}, see two recent reviews in \shortciteB{hamrick_role_2020,moerland_intersection_2021}).

This surveys focuses on goal-related mechanisms that are mostly orthogonal to the choice of underlying
optimization algorithm. In practice, however, most of the research in that space uses \drl methods.

\subsection{Solving the Multi-Goal RL Problem with Goal-Conditioned RL Algorithms}
Goal-conditioned agents see their behavior affected by the goal they pursue. This is formalized via goal-conditioned policies, that is policies which produce actions based on the environment state and the agent's current goal: $\Pi:\m{S}\times\m{Z}_\m{G}\to\m{A}$, where $\m{Z}_\m{G}$ is the space of goal embeddings corresponding to the goal space $\m{G}$ \shortcite{schaul2015universal}. Note that ensembles of policies can also be formalized this way, via a meta-policy $\Pi$ that retrieves the particular policy from a one-hot goal embedding $z_g$ \shortcite<\eg >{kaelbling1993learning,sutton2011horde}.

The idea of using a unique \rl agent to target multiple goals dates back to \shortciteA{kaelbling1993learning}. Later, the \horde architecture proposed to use interaction experience to update one value function per goal, effectively transferring to all goals the knowledge acquired while aiming at a particular one \shortcite{sutton2011horde}. In these approaches, one policy is trained for each of the goals and the data collected by one can be used to train others.

Building on these early results, \shortciteA{schaul2015universal} introduced \textit{Universal Value Function Approximators} (\uvfa). They proposed to learn a unique goal-conditioned value function and goal-conditioned policy to replace the set of value functions learned in \horde. Using neural networks as function approximators, they showed that \uvfas enable transfer between goals and demonstrate strong generalization to new goals.

The idea of \textit{hindsight learning} further improves knowledge transfer between goals \shortcite{kaelbling1993learning,andrychowicz2017hindsight}. Learning by hindsight, agents can reinterpret a past trajectory collected while pursuing a given goal in the light of a new goal. By asking themselves, \textit{what is the goal for which this trajectory is optimal?}, they can use the originally failed trajectory as an informative trajectory to learn about another goal, thus making the most out of every trajectory \shortcite{eysenbach2020rewriting}. This ability dramatically increases the sample efficiency of goal-conditioned algorithms and is arguably an important driver of the recent interest in goal-conditioned \rl approaches.

%%%%%%%%%%%%%%%%

\section{The Intrinsically Motivated Skills Acquisition Problem and the RL-IMGEP Framework}
\label{sec:im_pb_solution}
This section builds on the multi-goal \rl problem to formalize the \textit{intrinsically motivated skills acquisition problem}, in which goals are not externally provided to the agents but must be represented and generated by them (Section~\ref{sec:im_pb}). The following section discusses how to evaluate competency in such an open problem (Section~\ref{sec:eval}). Finally, we then propose an extension of the goal-conditioned \rl framework to tackle this problem: \textit{\textsl{rl}-based intrinsically motivated goal exploration process} framework (\rlimgep, Section~\ref{sec:gcimgep_solutions}). 

\subsection{The Intrinsically Motivated Skills Acquisition Problem}
\label{sec:im_pb}
In the \textit{intrinsically motivated skills acquisition problem}, the agent is set in an open-ended environment without any pre-defined goal and needs to acquire a repertoire of skills. Here, a skill is defined as the association of a goal embedding $z_g$ and the policy to reach it $\Pi_g$. A repertoire of skills is thus defined as the association of a repertoire of goals $\m{G}$ with a goal-conditioned policy trained to reach them $\Pi_\m{G}$. The intrinsically motivated skills acquisition problem can now be modeled by a reward-free \mdp $\m{M}\,=\,\{\m{S},\, \m{A},\, \m{T},\, \rho_0\}$ that only characterizes the agent, its environment and their possible interactions. Just like children, agents must be autotelic, \ie they should learn to represent, generate, pursue and master their own goals.

\subsection{Evaluating RL-IMGEP Agents}
\label{sec:eval}

Evaluating agents is often trivial in reinforcement learning. Agents are trained to maximize one or several pre-coded reward functions\,---\,the set of possible interactions is known in advance. One can measure generalization abilities by computing the agent's success rate on a held-out set of testing goals. One can measure exploration abilities via several metrics such as the count of task-specific state visitations.

In contrast, autotelic agents evolve in open-ended environments and learn to represent and form their own set of skills. In this context, the space of possible behaviors might quickly become intractable for the experimenter, which is perhaps the most interesting feature of such agents. For these reasons, designing evaluation protocols is not trivial.

The evaluation of such systems raises similar difficulties as the evaluation of task-agnostic content generation systems like Generative Adversarial Networks (\gan) \shortcite{goodfellow2014generative} or self-supervised language models \shortcite{devlin2019bert,brown2020language}. In both cases, learning is \textit{task-agnostic} and it is often hard to compare models in terms of their outputs (\eg comparing the quality of \gan output images, or comparing output repertoires of skills in autotelic agents).

One can also draw parallel with the debate on the evaluation of open-ended systems in the field of \textit{open-ended evolution} \shortcite{hintze_open-endedness_2019,stanley_role_2016,stanley_why_2019}. In both cases, a \textit{good} system is expected to generate more and more original solutions such that its output cannot be predicted in advance. But what does \textit{original} mean, precisely? \shortciteA{stanley_role_2016} argues that subjectivity has a role to play in the evaluation of open-ended systems. Indeed, the notion of \textit{interestingness} is tightly coupled with that of \textit{open-endedness}. What we expect from our open-ended systems, and of our \rlimgep agents in particular, is to generate more and more behaviors that \textit{we} deem interesting. This is probably why the evaluation of content generators often include human studies. Our end objective is to generate interesting artefacts for us; we thus need to evaluate open-ended processes ourselves, subjectively.

Our end goal would be to interact with trained \rlimgep directly, to set themselves goals and test their abilities. The evaluation would need to adapt to the agent's capabilities. As Einstein said \textit{``If you judge a fish by its ability to climb a tree, it will live its whole life believing that it is stupid.''}. \rlimgep need to be evaluated by humans looking for their area of expertise, assessing the width and depth of their capacities in the world they were trained in. This said, science also requires more objective evaluation metrics to facilitate the comparison of existing methods and enable progress. Let us list some evaluation methods measuring the competency of agents via proxies:

\begin{itemize}
    \item \textbf{Measuring exploration:} one can compute task-agnostic exploration proxies such as the entropy of the
    visited state distribution, or measures of state coverage (\eg coverage of the high-level x-y state space in mazes) \shortcite{goalgan}. Exploration can also be measured as the number of interactions from a set of \textit{interesting} interactions defined subjectively by the experimenter \shortcite<\eg interactions with objects in>{imagine}.
    \item \textbf{Measuring generalization:} The experimenter can subjectively define a set of relevant target goals
    and prevent the agent from training on them. Evaluating agents on this held-out set at test time provides a measure of generalization \shortcite{ruis2020benchmark}, although it is biased towards what the experimenter assesses as \textit{relevant} goals.
    \item \textbf{Measuring transfer learning:} The intrinsically motivated exploration of the environment can be seen as a pre-training phase to bootstrap learning in a subsequent downstream task. In the downstream task, the agent is trained to achieve externally-defined goals. We report its performance and learning speed on these goals. This is akin to the evaluation of self-supervised language models, where the reported metrics evaluate performance in various downstream tasks \shortcite<\eg >{brown2020language}. In this evaluation setup, autotelic agents can be compared to task-specific agents. Ideally, autotelic agents should benefit from their open-ended learning process to outperform task-specific agents on their own tasks. This said, performance on downstream tasks remains an evaluation proxy and should not be seen as the explicit \textit{objective} of the skill discovery phase. Indeed, in humans, skill discovery processes do not target any specific future task, but emerged from a natural evolutionary process maximizing reproductive success, see a discussion in \shortciteA{singh2010intrinsically}.
    \item
    \textbf{Opening the black-box:} Investigating internal representations learned during intrinsically motivated exploration is often informative. One can investigate properties of the goal generation system (\eg does it generate out-of-distribution goals?), investigate properties of the goal embeddings (\eg are they disentangled?). One can also look at the learning trajectories of the agents across learning, especially when they implement their own curriculum learning \shortcite<\eg >{goalgan,curious,blaes2019control,pong2019skew,akakzia2020decstr}.
    \item
    \textbf{Measuring robustness:} Autonomous learning agents evolving in open-ended environment should be robust to a variety of properties than can be found in the real-world. This includes very large environments, where possible interactions might vary in terms of difficulty (trivial interactions, impossible interactions, interactions whose result is stochastic thus prevent any learning progress). Environments can also include distractors (\eg non-controllable objects) and various forms of non-stationarity. Evaluating learning algorithms in various environments presenting each of these properties allows to assess their ability to solve the corresponding challenges.
\end{itemize}

%\newpage
\subsection{RL-Based Intrinsically Motivated Goal Exploration Processes}
\label{sec:gcimgep_solutions}
Until recently, the \imgep family was powered by population-based algorithms (\popimgep). The emergence of goal-conditioned \rl approaches that generate their own goals gave birth to a new type of \imgeps: the \rl-based \imgeps (\rlimgep). This section builds on traditional \rl and goal-conditioned \rl algorithms to give a general definition of intrinsically motivated goal-conditioned \rl algorithms (\rlimgep).

\begin{figure}[h]
    \centering
    \includegraphics[width=0.95\textwidth]{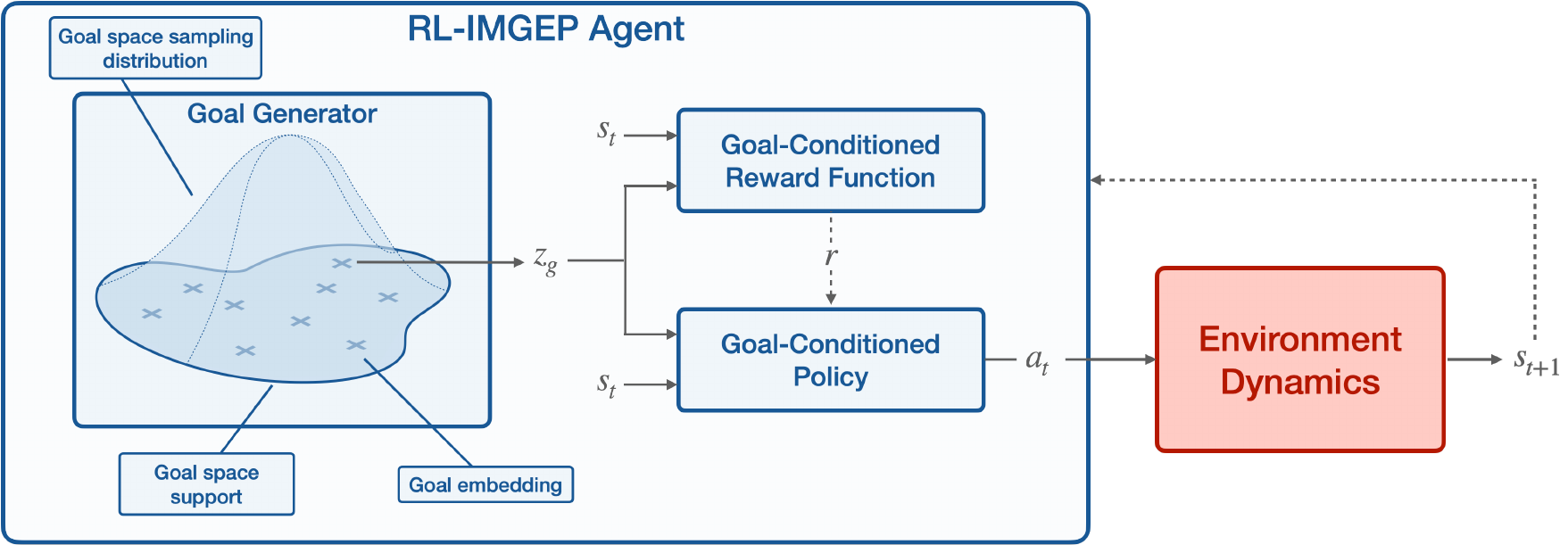}
    \caption{Representation of the different learning modules in a \rlimgep algorithm. In contrast, externally     motivated goal exploration processes (\rlemgeps) only train the goal-conditioned policy and assume     \textit{external} goal generator and goal-conditioned reward function. Learning goal embeddings, goal space
    support and goal-conditioned reward functions are all about learning to \textit{represent goals}. Learning a
    sampling distribution is about learning to \textit{prioritize their selection}.}
    \label{fig:goal_directed_rl}
\end{figure}

\rlimgep are intrinsically motivated versions of goal-conditioned \rl algorithms. They need to be equipped with mechanisms to represent and generate their own goals in order to solve the intrinsically motivated skills acquisition problem, see Figure~\ref{fig:goal_directed_rl}. Concretely, this means that, in addition to the goal-conditioned policy, they need to learn: 1)~to represent goals $g$ by compact embeddings $z_g$; 2)~to represent the support of the goal distribution, also called \textit{goal space} $\m{Z}_\m{G}=\{z_g\}_{g\in\m{G}}$; 3)~a goal distribution from which targeted goals are sampled $\m{D}(z_g)$; 4)~a goal-conditioned reward function $\m{R}_\m{G}$. In practice, only a few architectures tackle the four learning problems above. 

In this survey, \textbf{we call \textit{autotelic} any architecture where the agent selects its own goals (learning problem 3)}. Simple autotelic agents assume pre-defined goal representations (1), the support of the goals distribution (2) and goal-conditioned reward functions (4). As autotelic architectures tackle more of the 4 learning problems, they become more and more advanced. As we will see in the following sections, many existing works in goal-conditioned \rl can be formalized as autotelic agents by including goal sampling mechanisms \textit{within the definition of the agent}. 

With a developmental perspective, one can reinterpret existing work through the autotelic \rl framework. Let us take an example. The \textsc{agent$_{57}$} algorithm automatically selects a parameter to balance the intrinsic and extrinsic rewards of the agent at the beginning of each training episode \shortcite{badia2020agent57}. The authors do not mention the concept of \textit{goal} but instead present this mechanism as a form of reward shaping technique independent from the agent. With a developmental perspective, one can interpret the mixing parameter as a goal embedding. Replacing the sampling mechanism within the boundaries of the agent, \textsc{agent$_{57}$} becomes autotelic. It is intrinsically motivated to sample and target its own goals; \ie to define its own reward functions (here mixtures of intrinsic and extrinsic reward functions). 

Algorithm~\ref{algo:IMGEP} details the pseudo-code of \rlimgep algorithms. Starting from randomly initialized modules and memory, \rlimgep agents enter a standard \rl interaction loop. They first observe the context (initial state), then sample a goal from their goal sampling policy. Then starts the proper interaction. Conditioned on their current goal embedding, they act in the world so as to reach their goal, \ie to maximize the cumulative rewards generated by the goal-conditioned reward function. After the interaction, the agent can update all its internal models. It learns to represent goals by updating its goal embedding function and goal-conditioned reward function, and improves its behavior towards them by updating its goal-conditioned policy. 

This surveys focuses on the mechanisms specific to \rlimgep agents, \ie mechanisms that handle the representation, generation and selection of goals. These mechanisms are mostly orthogonal to the question of how to reach the goals themselves, which often relies on existing goal-conditioned algorithms, but can also be powered by imitation learning, evolutionary algorithms or other control and planning methods. Section~\ref{sec:survey_goal_rep} first presents a typology of goal representations used in the literature, before Sections~\ref{sec:survey_learning_goal_rep}~and~\ref{sec:survey_generation} cover existing methods to learn to represent and prioritize goals respectively.

\noindent\begin{minipage}{\textwidth}
   \centering
   \begin{minipage}{.6\textwidth}
        \begin{algorithm}[H]
            \small
        	\caption{~ Autotelic Agent with RL-IMGEP}
        	\label{algo:IMGEP}
        	\begin{algorithmic}[1]
            	\Require environment $\m{E}$
            	\State \textbf{Initialize} empty memory $\m{M}$,\\goal-conditioned policy $\Pi_\m{G}$, goal-conditioned reward $R_\m{G}$,\\goal space $\m{Z}_\m{G}$, goal sampling policy $GS$.
            	
            	\Loop
            	
                \LineCommentConttwo{\textit{Observe context}}
                \State Get initial state: $s_0 \leftarrow \m{E}$.reset()
            	\LineCommentCont{\textit{Sample goal}}
            	\State Sample goal embedding $z_g=GS(s_0, \m{Z}_\m{G})$.
            	\LineCommentCont{\textit{Roll-out goal-conditioned policy}}
            	\State Execute a roll-out with $\Pi_g=\Pi_\m{G}(\cdot \mid z_g)$
            	\State Store collected transitions $\tau=(s,a,s')$ in $\m{M}$.
            	\LineCommentCont{\textit{Update internal models}}
                \State Sample a batch of $B$ transitions: $\m{M}\sim \{(s,a,s')\}_B$.
                \State Perform Hindsight Relabelling $\{(s,a,s',z_g)\}_B$.
                \State Compute internal rewards $r=R_\m{G}(s,a,s'\mid z_g)$.
            	\State Update policy $\Pi_\m{G}$ via \rl on $\{(s,a,s',z_g,r)\}_B$.
            	\State Update goal representations  $\m{Z}_\m{G}$. 
            	\State Update goal-conditioned reward function $R_\m{G}$. 
            	\State Update goal sampling policy $GS$.
            	\EndLoop
            	\State \Return $\Pi_\m{G}, R_\m{G}, \m{Z}_\m{G}$
        	\end{algorithmic}
        \end{algorithm}
   \end{minipage}
   \hspace{0.2cm}
   \begin{minipage}{.37\textwidth}
  \textbf{ } \\
  \\
        \centering
        \includegraphics[width=0.85\textwidth]{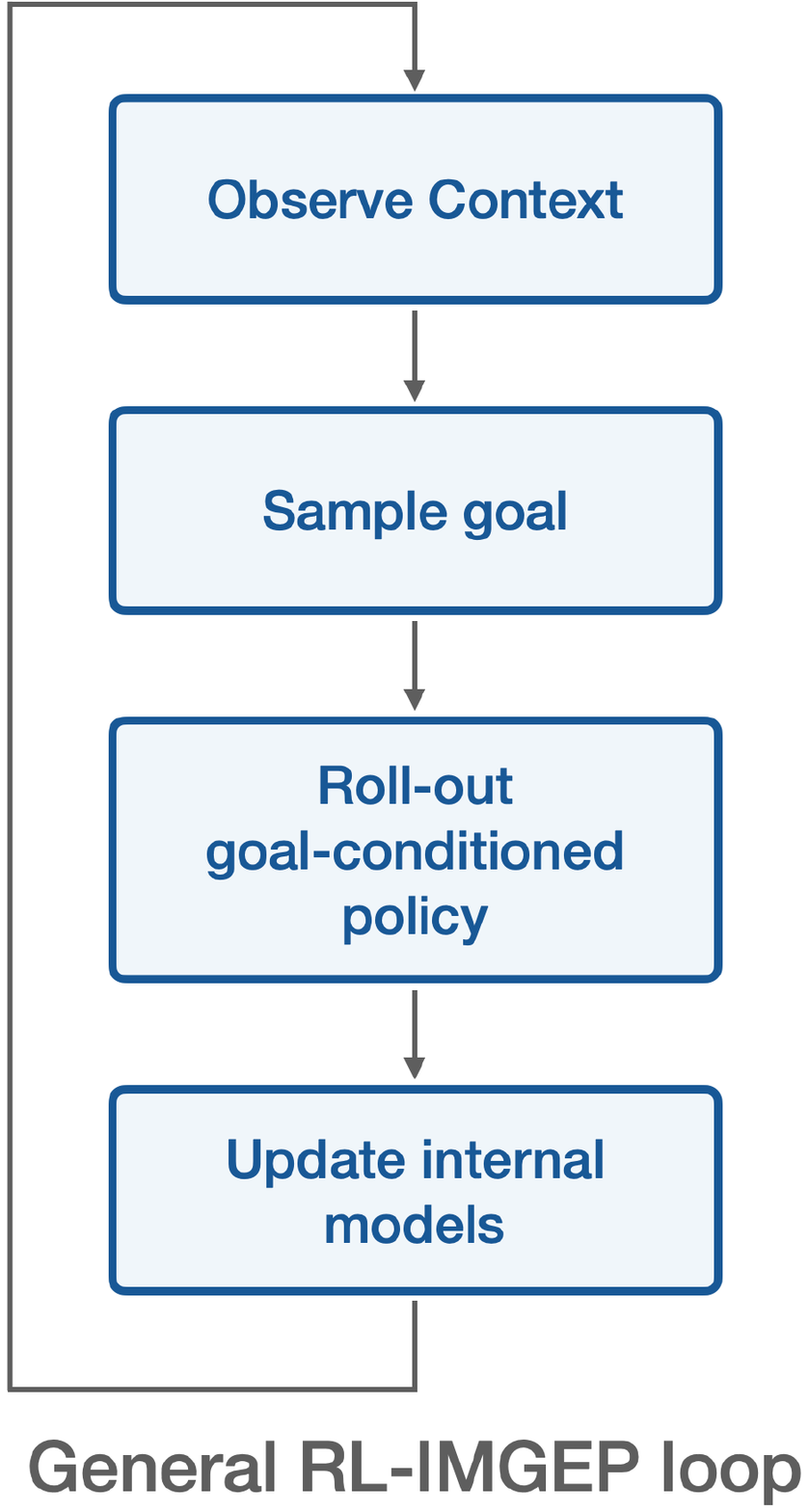}
    \end{minipage}
   \label{fig:test}
\end{minipage}

% \begin{algorithm}[H]
% \small
% 	\caption{~ Intrinsically Motivated Goal-Directed RL}
% 	\label{algo:IMGEP}
% 	\begin{algorithmic}[1]
% 	\Require environment $\m{E}$
% 	%\Require Environment $E$ with dynamics $\delta_E$
	
% 	\State \textbf{Initialize} empty memory $\m{M}$, goal-conditioned policy $\Pi_\m{G}$,\\goal-conditioned reward $R_\m{G}$, goal space $\m{Z}_\m{G}$,\\goal sampling policy $GS$.
	
% 	\Loop
	
%     \LineCommentConttwo{Data collection}
%     \State Get initial state: $s_0 \leftarrow \m{E}$.reset()
% 	\State Sample goal embedding $z_g=GS(s_0, \m{Z}_\m{G})$.
% 	\State Execute a roll-out with $\Pi_g=\Pi_\m{G}(\cdot \mid z_g)$, observe traj. $\tau$.
% 	\State Store collected transitions $\tau=(s,a,s')$ in $\m{M}$.
% 	\\
% 	\LineCommentCont{Data exploitation}
%     \State Sample a batch of $B$ transitions: $\m{M}\sim \{(s,a,s')\}_B$.
%     \State Perform Hindsight Relabelling $\{(s,a,s',z_g)\}_B$.
%     \State Compute internal rewards $r=R_\m{G}(s,a,s'\mid z_g)$.
% 	\State Update policy $\Pi$ via \rl on $\{(s,a,s',z_g,r)\}_B$.
% 	\State Update goal representations  $\m{Z}_\m{G}$. 
% 	\State Update goal-conditioned reward function $R_\m{G}$. 
% 	\State Update goal sampling policy $GS$.
% 	\EndLoop
% 	\State \Return $\Pi, R_\m{G}, \m{Z}_\m{G}$
% 	\end{algorithmic}
% \end{algorithm}

% % % % % % % % % % % % % % % % % % % % % % % % % % %
% A Survey of Goal Representations in the Literature
% % % % % % % % % % % % % % % % % % % % % % % % % % % 

\section{A Typology of Goal Representations in the Literature}
\label{sec:survey_goal_rep}

\begin{figure}[h]
    \centering
    \includegraphics[width=\textwidth]{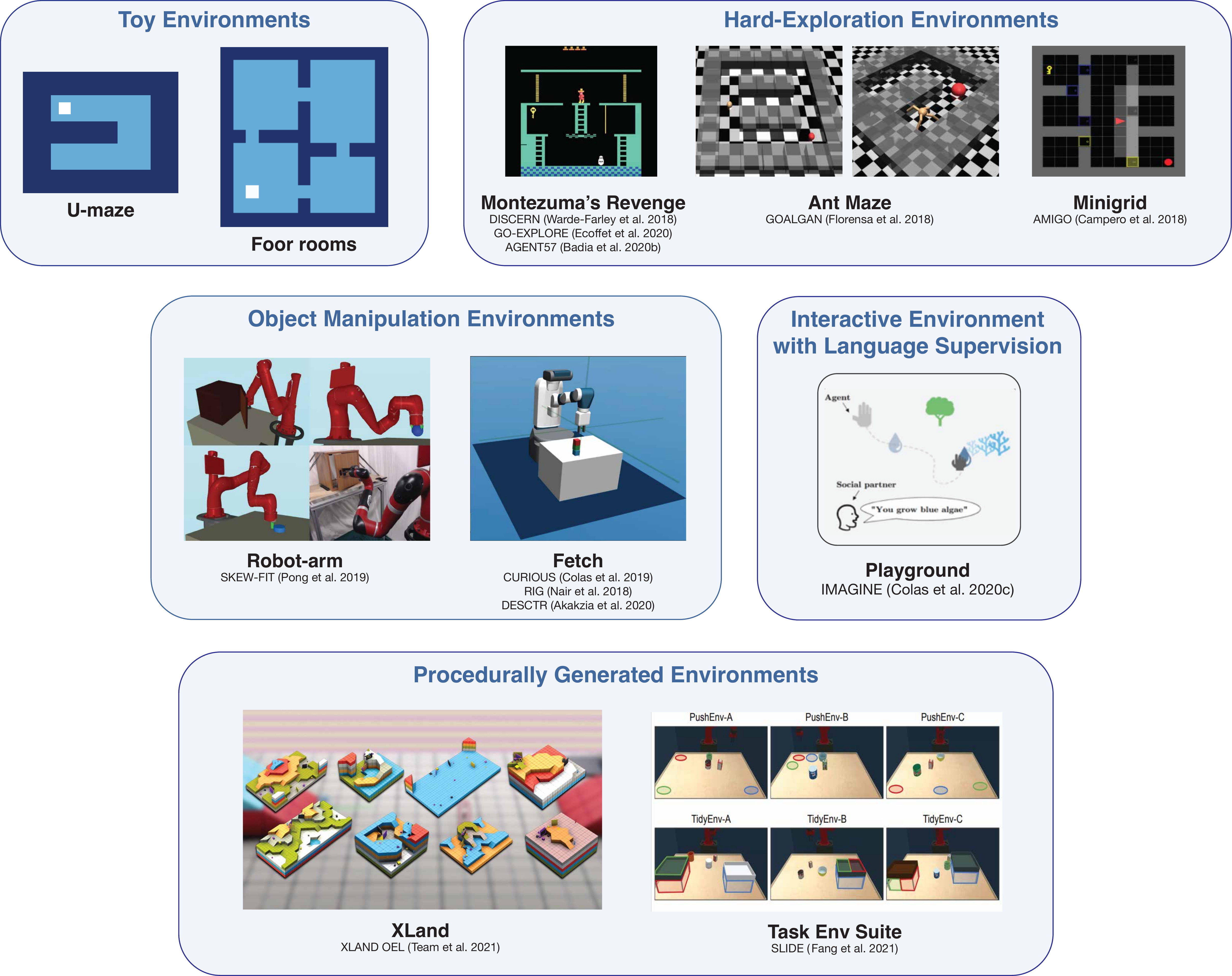}
    \caption{\textbf{Examples of environments in autotelic RL approaches.} We organize them by dominant feature but they might share features from other catagories as well. \textit{Toy Envs.} are used to investigate and visualise goal-as-state coverage over 2D worlds; \textit{Hard-Exploration Envs.} are used to benchmark goal generation algorithms; \textit{Object Manipulation Envs.} allow for the study of the diversity of learned goals as well as curriculum learning; \textit{Interactive Envs} permit to represent goals using language and to model interaction with caregivers; \textit{Procedurally Generated Envs.} enhance the vastness of potentially reachable goals.}
    \label{fig:envs}
\end{figure}

Now that we defined the problem of interest and the overall framework to tackle it, we can start reviewing relevant approaches from the literature and how they fit in this framework. This section presents a typology of the different kinds of goal representations found in the literature. Each goal is represented by a pair: 1) a \textit{goal embedding} and 2) a goal-conditioned reward function. Figure~\ref{fig:envs} also provides visuals of the main environments used by the autotelic approaches presented in this paper.

\subsection{Goals as Choices Between Multiple Objectives}
Goals can be expressed as a list of different objectives the agent can choose from.

\paragraph{Goal embedding.} In that case, goal embeddings $z_g$ are one-hot encodings of the current objective being pursued among the $N$ objectives available. $z_g^i$ is the $i^\text{th}$ one-hot vector: $z_g^i\,=\,(\mathds{1}_{j=i})_{j=[1..N]}$. This is the case in \shortciteA{oh2017zero,mankowitz2018unicorn,codevilla2018end}.

\paragraph{Reward function.} The goal-conditioned reward function is a collection of $N$ distinct reward functions $R_\m{G}(\cdot)=R_i(\cdot)$ if $z_g=z_g^i$. In \shortciteA{mankowitz2018unicorn} and \shortciteA{chan2019actrce}, each reward function gives a positive reward when the agent reaches the corresponding object: reaching guitars and keys in the first case, monsters and torches in the second.

\subsection{Goals as Target Features of States}
Goals can be expressed as target features of the state the agent desires to achieve.

\paragraph{Goal embedding.} In this scenario, a state representation function $\varphi$ maps the state space to an embedding space $\m{Z}=\varphi(\m{S})$. Goal embeddings $z_g$ are target points in $\m{Z}$ that the agent should reach. In manipulation tasks, $z_g$ can be target block coordinates \shortcite{andrychowicz2017hindsight,nair2017overcoming,plappert2018multi,curious,fournier2019clic,blaes2019control,lanier2019curiosity,ding_imitation_2019,li2019towards}. In navigation tasks, $z_g$ can be target agent positions \shortcite<\eg in mazes, >{schaul2015universal,goalgan}. Agent can also target image-based goals. In that case, the state representation function $\varphi$ is usually implemented by a generative model trained on experienced image-based states and goal embeddings can be sampled from the generative model or encoded from real images \shortcite{zhu2017target,codevilla2018end,nair2018visual,pong2019skew,warde2018unsupervised,florensa2019selfsupervised,venkattaramanujam2019self,pmlr-v100-lynch20a,lynch2020grounding,nair2020contextual,kovac2020grimgep}.

\paragraph{Reward function.} For this type of goals, the reward function $R_\m{G}$ is based on a distance metric $D$. One can define a dense reward as inversely proportional to the distance between features of the current state and the target goal embedding: $R_g=R_\m{G}(s|z_g)=-\alpha\times D(\varphi(s),~z_g)$ \shortcite<e.g. >{nair2018visual}. The reward can also be sparse: positive whenever that distance falls below a pre-defined threshold: $R_\m{G}(s|z_g)\,=\,1~\text{if}~D(\varphi(s),\,z_g)<\epsilon$, $0$ otherwise.

\subsection{Goals as Abstract Binary Problems}
Some goals cannot be expressed as target state features but can be represented by \textit{binary problems}, where each goal expresses as set of constraint on the state (or trajectory) such that these constraints are either verified or not (binary goal achievement).

\paragraph{Goal embeddings.} In binary problems, goal embeddings can be any expression of the set of constraints
that the state should respect. \shortciteA{akakzia2020decstr,ecoffet2020first} both propose a pre-defined discrete state representation. These representations lie in a finite embedding space so that goal completion can be asserted when the current embedding $\varphi(s)$ equals the goal embedding $z_g$. Another way to express sets of constraints is via language-based predicates. A sentence describes the constraints expressed by the goal and the state or trajectory
either verifies them, or does not \shortcite{Hermann2017,chan2019actrce,Jiang2019,bahdanau2018learning,bahdanau2018systematic,hill2019emergent,ther,imagine,lynch2020grounding}, see \shortcite{Luketina2019} for a recent review. Language can easily characterize \textit{generic goals} such as ``\textit{grow any blue object}'' \shortcite{imagine}, \textit{relational goals} like ``\textit{sort objects by size}" \shortcite{Jiang2019}, ``\textit{put the cylinder in the drawer}" \shortcite{lynch2020grounding} or even \textit{sequential goals} ``\textit{Open the yellow door after you open a purple door}'' \shortcite{chevalier-boisvert2018babyai}. When goals can be expressed by language sentences, goal embeddings $z_g$ are usually language embeddings learned jointly with either the policy or the reward function. Note that, although \rl goals always express constraints on the state, we can imagine \textit{time-extended goals} where constraints are expressed on the trajectory (see a discussion in Section~\ref{sec:future_diversity}).

\paragraph{Reward function.} The reward function of a binary problem can be viewed as a binary classifier that evaluates whether state $s$ (or trajectory $\tau$) verifies the constraints expressed by the goal semantics (positive reward) or not (null reward). This binary classification setting has directly been implemented as a way to learn language-based goal-conditioned reward functions $R_g(s\mid z_g)$ in \shortciteA{bahdanau2018learning} and \shortciteA{imagine}. Alternatively, the setup described in \shortciteA{colas2020language} proposes to turn binary problems expressed by language-based goals into goals as specific target features. To this end, they train a language-conditioned goal generator that produces specific target features verifying constraints expressed by the binary problem. As a result, this setup can use a distance-based metric to evaluate the fulfillment of a binary goal.

\subsection{Goals as a Multi-Objective Balance}
Some goals can be expressed, not as desired regions of the state or trajectory space but as more general objectives that the agent should maximize. In that case, goals can parameterize a particular mixture of multiple objectives that the agent should maximize.

\paragraph{Goal embeddings.} Here, goal embeddings are simply sets of weights balancing the different objectives $z_g\,=\,(\beta_i)_{i=[1..N]}$ where $\beta_i$ is the weights applied to objective $i$ and $N$ is the number of objectives. Note that, when $\beta_j\,=\,1$ and $\beta_i\,=v0,~\forall i\neq j$, the agent can decide to pursue any of the objective alone. In \textit{Never Give Up}, for example, \rl agents are trained to maximize a mixture of extrinsic and intrinsic rewards \shortcite{badia2020never}. The agent can select the mixing parameter $\beta$ that can be viewed as a goal. Building on this approach, \textsc{agent$_{57}$} adds a control of the discount factor, effectively controlling the rate at which rewards are discounted as time goes by \shortcite{badia2020agent57}.

\paragraph{Reward function.} When goals are represented as a balance between multiple objectives, the associated reward function cannot be represented neither as a distance metric, nor as a binary classifier. Instead, the agent needs to maximize a convex combination of the objectives: $R_g(s)\,=\,\sum_{i=1}^N~\beta_g^i R^i(s)$ where $R^i$ is the $i^\text{th}$ of $N$ objectives and $z_g=\beta=\beta_i^g\mid_{i\in[1..N]}$ is the set of weights.

\subsection{Goal-Conditioning}
Now that we described the different types of goal embeddings found in the literature, remains the question of how to condition the agent's behavior\,---\,\ie the policy\,---\,on them. Originally, the \uvfa framework proposed to concatenate the goal embedding to the state representation to form the policy input. Recently, other mechanisms have emerged. When language-based goals were introduced, \shortciteA{chaplot2017gatedattention} proposed the \textit{gated-attention} mechanism where the state features are linearly scaled by attention coefficients computed from the goal representation $\varphi(z_g)$: $\text{input}\,=\,s\,\odot\,\varphi(z_g)$, where $\odot$ is the Hadamard product. Later, the Feature-wise Linear Modulation (\textsc{film}) approach \shortcite{perez2018film} generalized this principle to affine transformations: $\text{input}\,=\,s\,\odot\,\varphi(z_g)\, +\,\psi(z_g)$. Alternatively, \shortciteA{Andreas_2016} came up with \textit{Neural Module Networks}, a mechanism that leverages the linguistic structure of goals to derive a symbolic program that defines how states should be processed \shortcite{bahdanau2018learning}.

\subsection{Conclusion}
This section presented a diversity of goal representations, corresponding to a diversity of reward functions architectures. However, we believe this represents only a small fraction of the diversity of goal types that humans pursue. Section~\ref{sec:future} discusses other goal representations that \rl algorithms could target.

% % % % % % % % % % % % % % % % % % % % % % % % % % %
% How to learn goal representations?
% % % % % % % % % % % % % % % % % % % % % % % % % % % 

\section{How to Learn Goal Representations?}
\label{sec:survey_learning_goal_rep}

The previous section discussed various types of goal representations. Autotelic agents actually need to learn these goal representations. While individual goals are represented by their embeddings and associated reward functions, representing multiple goals also requires the representation of the \textit{support} of the goal space, \ie how to represent the collection of \textit{valid goals} that the agent can sample from, see Figure~\ref{fig:goal_directed_rl}. This section reviews different approaches from the literature.

\subsection{Assuming Pre-Defined Goal Representation}
Most approaches tackle the multi-goal \rl problem, where goal spaces and associated rewards are pre-defined by the engineer and are part of the task definition. Navigation and manipulation tasks, for example, pre-define goal spaces (\eg target agent position and target block positions respectively) and use the Euclidean distance to compute rewards \shortcite{schaul2015universal,andrychowicz2017hindsight,nair2017overcoming,plappert2018multi,goalgan,curious,blaes2019control,lanier2019curiosity,ding_imitation_2019,li2019towards}. \shortciteA{akakzia2020decstr,ecoffet2020first} hand-define abstract state representation and provide positive rewards when these match target goal representations. Finally, \shortciteA{team2021open} hand-define a large combinatorial goal space, where goals are Boolean formulas of predicates such as \textit{being near, on, seeing,} and \textit{holding}, as well as their negations, with arguments taken as entities such as \textit{objects, players}, and \textit{floors} in procedurally-generated multi-player worlds.
In all these works, goals can only be sampled from a pre-defined bounded space. This falls short of solving the intrinsically motivated skills acquisition problem. The next sub-section investigates how goal representations can be learned.

\subsection{Learning Goal Embeddings}
Some approaches assume the pre-existence of a goal-conditioned reward function, but learn to represent goals by learning goal embeddings. This is the case of language-based approaches, which receive rewards from the environment (thus are \rlemgep), but learn goal embeddings jointly with the policy during policy learning \shortcite{Hermann2017,chan2019actrce,Jiang2019,bahdanau2018systematic,hill2019emergent,ther,lynch2020grounding}. When goals are target images, goal embeddings can be learned via generative models of states, assuming the reward to be a fixed distance metric computed in the embedding space \shortcite{nair2018visual,florensa2019selfsupervised,pong2019skew,nair2020contextual}.

\subsection{Learning the Reward Function}
\label{sec:survey_learning_goal_rep_rew}

A few approaches go even further and learn their own goal-conditioned reward function. \shortciteA{bahdanau2018learning,imagine} learn language-conditioned reward functions from an expert dataset or from language descriptions of autonomous exploratory trajectories respectively. However, the \textsc{agile} approach from \shortciteA{bahdanau2018learning} does not generate its own goals.

In the domain of image-based goals, \shortciteA{venkattaramanujam2019self,hartikainen2019dynamical} learn a distance metric estimating the square root of the number of steps required to move from any state $s_1$ to any $s_2$ and generates internal signals to reward agents for getting closer to their target goals. \shortciteA{warde2018unsupervised} learn a similarity metric in the space of controllable aspects of the environment that is based on a mutual information objective between the state and the goal state $s_g$.

\shortciteA{wu2018laplacian} compute a distance metric representing the ability of the agent to reach one state from another using the Laplacian of the transition dynamics graph, where nodes are states and edges are actions. More precisely, they use the eigenvectors of the Laplacian matrix of the graph given by the states of the environment as basis to compute the L2 distance towards a goal configuration.

Another way to learn reward function and their associated skills is via \textit{empowerment} methods \shortcite{mohamed_variational_2015,gregor2016variational,achiam_variational_2018,eysenbach2018diversity,dai_empowerment-based_2020,sharma_dynamics-aware_2020,choi_variational_2021}. Empowerment methods aim at maximizing the mutual information between the agent's actions or goals and its experienced states. Recent methods train agents to develop a set of skills leading to maximally different areas of the state space. Agents are rewarded for experiencing states that are easy to discriminate, while a discriminator is trained to better infer the skill $z_g$ from the visited states. This discriminator acts as a skill-specific reward function. 

All these methods set their own goals and learn their own goal-conditioned reward function. For these reasons, they can be considered as complete autotelic \rl algorithms.

\subsection{Learning the Support of the Goal Distribution}
The previous sections reviewed several approaches to learn goal embeddings and reward functions. To represent collections of goals, one also needs to represent the support of the goal distribution\,---\,which embeddings correspond to valid goals and which do not.

Most approaches consider a pre-defined, bounded goal space in which any point is a valid goal (\eg target positions within the boundaries of a maze, target block positions within the gripper's reach) \shortcite{schaul2015universal,andrychowicz2017hindsight,nair2017overcoming,plappert2018multi,curious,blaes2019control,lanier2019curiosity,ding_imitation_2019,li2019towards}. However, not all approaches assume pre-defined goal spaces. 

The \textit{option framework} \shortcite{sutton_between_1999,precup_temporal_2000} proposes to train a high-level policy to compose sequences of behaviors originating from learned low-level policies called \textit{options}. Each option can be seen as a goal-directed policy where the goal embedding is represented by its index in the set of options. When options are policies aiming at specific states, \textit{option discovery} methods learn the support of the goal space; they learn which goal--state are most useful to organize higher-level behaviors. \textit{Bottleneck states} are often targeted as good sub-goals. \shortciteA{mcgovern_automatic_2001} propose to detect states that are common to multiple successful trajectories. \shortciteA{simsek_using_2004} propose to select state with maximal relative novelty, i.e. when the average novelty of following states is higher than the average novelty of previous ones. \shortciteA{simsek_skill_2008} propose to leverage measures from graph theory.

The option-critic framework then opened the way to a wealth of new approaches \shortcite{bacon2017option}. Among those, methods based on \textit{successor features} \shortcite{NIPS2017_350db081,barreto2020fast,ramesh_successor_2019} propose to learn the option space using reward embeddings. With successor features, the Q-value of a goal can be expressed as a linear combination of learned reward features, efficiently decoupling the rewards from the environmental dynamics. In a multi-goal setting, these methods pair each goal with a reward embedding and use \textit{generalized policy improvement} to train a set of policies that efficiently share relevant reward features across goals. These methods provide key mechanisms to learn to discover and represent sub-goals. However, they do not belong to the \rlimgep family since high-level goals are externally provided.

Some approaches use the set of previously experienced representations to form the support of the goal distribution \shortcite{veeriah2018many,akakzia2020decstr,ecoffet2020first}. In \shortciteA{goalgan}, a Generative Adversarial Network (\gan) is trained on past representations of states ($\varphi(s)$) to model a distribution of goals and thus its support. In the same vein, approaches handling image-based goals usually train a generative model of image states based on Variational Auto-Encoders (\vae) to model goal distributions and support \shortcite{nair2018visual,pong2019skew,nair2020contextual}. In both cases, valid goals are the one generated by the generative model.

We saw that the support of valid goals can be pre-defined, a simple set of past representations or approximated by a generative model trained on these. In all cases, the agent can only sample goals \textit{within} the convex hull of previously encountered goals (in representation space). We say that goals are \textit{within} training distribution. This drastically limits exploration and the discovery of new behaviors.

Children, on the other hand, can imagine creative goals. Pursuing these goals is thought to be the main driver of exploratory play in children \shortcite{chu2020exploratory}. This is made possible by the compositionality of language, where sentences can easily be combined to generate new ones. The \imagine algorithm leverages the creative power of language to generate such \textit{out-of-distribution} goals \shortcite{imagine}. The support of valid goals is extended to any combination of language-based goals experienced during training. They show that this mechanism augments the generalization and exploration abilities of learning agents.

In Section~\ref{sec:survey_generation}, we discuss how agents can learn to adapt the goal sampling distribution
to maximize the learning progress of the agent.

\subsection{Conclusion}
This section presented how previous approaches tackled the problem of learning goal representations. While most approaches rely on pre-defined goal embeddings and/or reward functions, some approaches proposed to learn internal reward functions and goal embeddings jointly.

% % % % % % % % % % % % % % % % % % % % % % % % % % %
% How to generate goals?
% % % % % % % % % % % % % % % % % % % % % % % % % % % 

\section{How to Prioritize Goal Selection?}
\label{sec:survey_generation}

Autotelic agents also need to select their own goals. While goals can be generated by uninformed sampling of the goal space, agents can benefit from mechanisms optimizing goal selection. In practice, this boils down to the automatic adaptation of the goal sampling distribution as a function of the agent performance.

\subsection{Automatic Curriculum Learning for Goal Selection}
In real-world scenarios, goal spaces can be too large for the agent to master all goals in its lifetime. Some goals might be trivial, others impossible. Some goals might be reached by chance sometimes, although the agent cannot make any progress on them. Some goals might be reachable only after the agent mastered more basic skills. For all these reasons, it is important to endow autotelic agents learning in open-ended scenarios with the ability to optimize their goal selection mechanism. This ability is a particular case of \textit{automatic curriculum learning} \textsc{acl} applied for goal selection: mechanisms that organize goal sampling so as to maximize the long-term performance improvement (distal objective). As this objective is usually not directly differentiable, curriculum learning techniques usually rely on a proximal objective. In this section, we look at various proximal objectives used in automatic curriculum learning strategies to organize goal selection. Interested readers can refer to \shortciteA{portelas2020automatic}, which present a broader review of \textsc{acl} methods for \rl. Note that knowledge-based \ims can rely on similar proxies but focus on the optimization of the experienced states instead of on the selection of goals (\eg maximize next-state prediction errors). A recent review of knowledge-based \im approaches can be found in \shortciteA{linke2019adapting}.

\paragraph{Intermediate or uniform difficulty.} Intermediate difficulty has been used as a proxy for long-term performance improvement, following the intuition that focusing on goals of intermediate difficulty results in short-term learning progress that will eventually turn into long-term performance increase. \textsc{goalgan} assigns feasibility scores to goals as the proportion of time the agents successfully reaches it \shortcite{goalgan}. Based on this data, a \gan is trained to generate goals of intermediate difficulty, whose feasibility scores are contained within an intermediate range. \shortciteA{sukhbaatar2017intrinsic} and \shortciteA{campero2020learning} train a goal policy with \rl to propose challenging goals to the \rl agent. The goal policy is rewarded for setting goals that are neither too easy nor impossible. In the same spirit, \shortciteA{team2021open} use a mixture of three criteria to filter valid goals: 1) the agent has a low probability of scoring high; 2) the agent has a high probability of scoring higher than a control policy; 3) the control policy performs poorly. Finally, \shortciteA{zhang2020automatic} select goals that maximize the disagreement in an ensemble of value functions. Value functions agree when the goals are too easy (the agent is always successful) or too hard (the agent always fails) but disagree for goals of intermediate difficulty.

\shortciteA{settersolver} propose a variant of the \textsc{goalgan} approach and train a goal generator to sample goals of all levels of difficulty, uniformly. This approach seems to lead to better stability and improved performance on more complex tasks compared to \textsc{goalgan} \shortcite{goalgan}.

Note that measures of intermediate difficulty are sensitive to the presence of stochasticity in the environment. Indeed, goals of intermediate difficulty can be detected as such either because the agent has not yet mastered them, or because the environment makes them impossible to achieve sometimes. In the second case, the agent should not focus on them, because it cannot learn anything new. Estimating medium-term learning progress helps overcoming this problem (see below).

\paragraph{Novelty - diversity.}  \shortciteA{warde2018unsupervised,pong2019skew,pitis2020maximum} all bias the selection of goals towards sparse areas of the goal space. For this purpose, they train density models in the goal space. While \shortciteA{warde2018unsupervised,pong2019skew} aim at a uniform coverage of the goal space (diversity), \shortciteA{pitis2020maximum} skew the distribution of selected goals even more, effectively maximizing novelty. \shortciteA{kovac2020grimgep} proposed to enhance these methods with a goal sampling prior focusing goal selection towards controllable areas of the goal space. Finally, \shortciteA{Fang-RSS-21} use procedural content generation (\textsc{pcg}) to train a task generator that produces diverse environments in which agents can explore customized skills. 

These algorithms have strong connections with empowerment methods \shortcite{mohamed_variational_2015,gregor2016variational,achiam_variational_2018,eysenbach2018diversity,campos_explore_2020,sharma_dynamics-aware_2020,choi_variational_2021}. Indeed, the mutual information between goals and states that empowerment methods aim to maximize can be rewritten as: 
\begin{equation*}
    I(Z,\,S) = H(Z) - H(Z\mid S).
\end{equation*}
Thus, maximizing empowerment can be seen as maximizing the entropy of the goal distribution while minimizing the entropy of goals given experienced states. Algorithm that both learn to sample diverse goals $(H(Z)\nearrow)$ and learn to represent goals with variational auto-encoders $(H(Z|S)\searrow)$ can be seen as maximizing empowerment. The recent wealth of \textit{empowerment} methods, however, rarely discusses the link with autotelic agents: they do not mention the notion of goals or goal-conditioned reward functions and do not discuss the problem of goal representations \shortcite{gregor2016variational,achiam_variational_2018,eysenbach2018diversity,campos_explore_2020,sharma_dynamics-aware_2020}. In a recent paper, \shortciteA{choi_variational_2021} investigated these links and formalized a continuum of methods from empowerment to visual goal-conditioned approaches. 

While \textit{novelty} refers to the \textit{originality} of a reached outcome, \textit{diversity} is a term that can only be applied to a collection of these outcomes. An outcome will be said novel if it is semantically different from what exists in the set of known outcomes. A set of outcomes will be said \textit{diverse} when outcomes are far from each other and \textit{cover well} the space of possible outcomes. Note that agents can also express diversity in their behavior towards a unique outcome, a skill known as \textit{versatility} \shortcite{hausman_learning_2018,kumar_one_2020,osa_discovering_2021,celik_specializing_2021}.

\paragraph{Medium-term learning progress.} 
The idea of using learning progress (\lp) as a intrinsic motivation for artificial agents dates back to the 1990s \shortcite{schmidhuber1991curious,schmidhuber1991learning,kaplan_maximizing_2004,oudeyer_intrinsic_2007}. At that time, however, it was used as a knowledge-based \im and rewarded progress in predictions. From 2007, \shortcite{oudeyer2007intrinsic} suggested to use it as a competence-based \im to reward progress in competence instead. In such approaches, agents estimate their \lp in different regions of the goal space and bias goal sampling towards areas of high absolute learning progress using bandit algorithms \shortcite{baranes2013active,moulinfriergmm,forestier2016modular,fournier2018accuracy,fournier2019clic,curious,blaes2019control,portelas_teacher_2020,akakzia2020decstr}. Such estimations attempts to disambiguate the incompetency or uncertainty the agent could resolve with more practice (epistemic) from the one it could not (aleatoric). Agents should indeed focus on goals towards which they can make progress and avoid goals that are either too easy, currently too hard, or impossible. 

\shortciteA{forestier2016modular,curious,blaes2019control} and \shortciteA{akakzia2020decstr} organize goals into modules and compute average \lp measures over modules. \shortciteA{fournier2018accuracy} defines goals as a discrete set of precision requirements in a reaching task and computes \lp for each requirement value. The use of absolute \lp enables agents to focus back on goals for which performance decreases (due to perturbations or forgetting). \shortciteA{akakzia2020decstr} introduces the success rate in the value optimized by the bandit: $v\,=\,(1\,-\,\text{\textsc{sr}})\,\times\,\text{\textsc{lp}}$, so that agents favor goals with high absolute \lp and low competence.

\subsection{Hierarchical Reinforcement Learning for Goal Sequencing.}

Hierarchical reinforcement learning (\hrl) can be used to guide the sequencing of goals \shortcite{dayan1993feudal,sutton1998intra,sutton_between_1999,precup2001temporal}. In \hrl, a high-level policy is trained via \rl or planning to generate sequence of goals for a lower level policy so as to maximize a higher-level reward. This allows to decompose tasks with long-term dependencies into simpler sub-tasks. Low-level policies are implemented by traditional goal-conditioned \rl algorithms \shortcite{levy2018hierarchical,roder2020curious} and can be trained independently from the high-level policy \shortcite{kulkarni2016hierarchical,frans2017meta} or jointly \shortcite{levy2018hierarchical,nachum2018data,roder2020curious}. In the option framework, option can be seen as goal-directed policies that the high-level policy can choose from~\shortcite{sutton_between_1999,precup_temporal_2000}. In that case, goal embeddings are simple indicators. Most approaches consider hand-defined spaces for the sub-goals (\eg positions in a maze). Recent approaches propose to use the state space directly \shortcite{nachum2018data} or to learn the sub-goal space (e.g. \shortciteA{vezhnevets2017feudal}, or with generative model of image states in  \shortciteA{nasiriany2019planning}).

% % % % % % % % % % % % % % % % % % % % % % % % % % %
% Future avenues
% % % % % % % % % % % % % % % % % % % % % % % % % % % 

\section{Open Challenges}
\label{sec:future}
This section discusses open challenges in the quest for autotelic agents tackling the intrinsically motivated skills acquisition problem. 

\subsection{Challenge \#1: Targeting a Greater Diversity of Goals}
\label{sec:future_diversity}
Section~\ref{sec:survey_goal_rep} introduces a typology of goal representations found in the literature. The diversity of goal representations seems however limited, compared to the diversity of goals human target \shortcite{ram1995goal}.

\paragraph{Time-extended goals.} All \rl approaches reviewed in this paper consider \textit{time-specific} goals, that is, goals whose completion can be assessed from any state $s$. This is due to the Markov property requirement, where the next state and reward need to be a function of the previous state only. \textit{Time-extended} goals\,---\,\ie goals whose completion can be judged by observing a sequence of states (\eg \textit{jump twice})\,---\,can however be considered by adding time-extended features to the state \shortcite<\eg the difference between the current state and the initial state>{imagine}. To avoid such \textit{ad-hoc} state representations, one could imagine using reward function architectures that incorporate forms of memory such as Recurrent Neural Network (\rnn) architectures \shortcite{elman} or Transformers \shortcite{vaswani2017attention}. Although recurrent policies are often used in the literature \shortcite{chevalier-boisvert2018babyai,hill2019emergent,loynd2019working,goyal2019recurrent}, recurrent reward functions have not been much investigated. Some work \shortciteA{sutton_tdnet_2004,schlegel_general_2021} investigate the benefit of computing relations between value functions when learning predictive representations. \shortciteA{sutton_tdnet_2004} propose to represent the interrelation of predictions in a \textit{TD-network} where nodes are predictions computed from states. The network allows to perform predictions that have complex temporal semantics. \shortciteA{schlegel_general_2021} train a RNN architecture where hidden-states are multi-step predictions. Finally, recent work by \shortciteA{karch2021grounding} show that agents can derive rewards from linguistic descriptions of time-extended behaviors.  Time-extended goals include interactions that span over multiple time steps (\eg \textit{shake the blue ball}) and spatio-temporal references to objects (\eg \textit{get the red ball that was on the left of the sofa yesterday}).

%with the exception of the recent work by . Time-extended goals include goals expressed as repetitions of a given interaction (\eg \textit{knock three times}) and sequential goals (\eg \textit{get the blue ball that was on the table yesterday, then roll it towards me}).

\paragraph{Learning goals.}
\textit{Goal-driven learning} is the idea that humans use \textit{learning goals}, goals about their own learning abilities as a way to simplify the realization of \textit{task goals} \shortcite{ram1995goal}. Here, we refer to \textit{task goals} as goals that express constraints on the physical state of the agent and/or environment. On the other hand, \textit{learning goals} refer to goals that express constraints on the knowledge of the agent. Although most \rl approaches target task goals, one could envision the use of \textit{learning goals} for \rl agents. 

In a way, learning-progress-based learning is a form of learning goal: as the agent favors regions of the goal space to sample its task goals, it formulates the goal of learning about this specific goal region \shortcite{baranes2013active,fournier2018accuracy,fournier2019clic,curious,blaes2019control,akakzia2020decstr}.

Embodied Question Answering problems can also be seen as using learning goals. The agent is asked a question (\ie a learning goal) and needs to explore the environment to answer it (acquire new knowledge) \shortcite{das2018embodied,Yuan_2019}. 

In the future, one could envision agents that set their own learning targets as sub-goals towards the resolution of harder task or learning goals, \eg \textit{I'm going to learn about knitting so I can knit a pullover to my friend for his birthday.}
%\os{Meta-Learning is generally defined as learning to learn. For the agent, learning the reward function or having learning goals and learning to reach them is about learning to learn. Maybe this general statement and a few refs should come earlier.}

\paragraph{Goals as optimization under selected constraints.}
We discussed the representations of goals as a balance between multiple objectives. An extension of this idea is to integrate the selection of constraints on states or trajectories. One might want to maximize a given metric (\eg walking speed), while setting various constraints (\eg maintaining the power consumption below a given threshold or controlling only half of the motors). The agent could explore in the space of constraints, setting constraints to itself, building a curriculum on these, etc. This is partially investigated in \shortciteA{colas2020epidemioptim}, where the agent samples constraint-based goals in the optimization of control strategies to mitigate the economic and health costs in simulated epidemics. This approach, however, only considers constraints on minimal values for the objectives and requires the training of an additional Q-function per constraint.

\paragraph{Meta-diversity of goals.} Finally, autotelic agents should learn to target all these goals within the same run; to transfer their skills and knowledge between different types of goals. For instance, targeting visual goals could help the agent explore the environment and solve learning goals or linguistic goals. As the density of possible goals increases, agents can organize more interesting curricula. They can select goals in easier representation spaces first (\eg sensorimotor spaces), then move on to target more difficult goals (\eg in the visual space), before they can target the more abstract goals (\eg learning goals, abstract linguistic goals). 

This can take the form of goal spaces organized hierarchically at different levels of abstractions. The exploration of such complex goal spaces has been called \textit{meta-diversity}  \shortcite{etcheverry_hierarchically-organized_2020}. In the outer-loop of the meta-diversity search, one aims at learning a diverse set of outcome/goal representations. In the inner-loop, the exploration mechanism aims at generating a diversity of behaviors in each existing goal space. How to efficiently transfer knowledge and skills between these multi-modal goal spaces and how to efficiently organize goal selection in large multi-modal goal spaces remains an open question.

\subsection{Challenge \#2: Learning to Represent Diverse Goals}
\label{sec:future_imagination_and_composition}
This survey mentioned only a handful of complete autotelic architectures. Indeed, most of the surveyed approach assume pre-existing goal embeddings or reward functions. Among the approaches that learn goal representations autonomously, we find that the learned representations are often restricted to very specific domains. Visual goal-conditioned approaches for example, learn reward functions and goal embeddings but restrict them to the visual space \shortcite{nair2018visual,nair2020contextual,warde2018unsupervised,venkattaramanujam2019self,pong2019skew,hartikainen2019dynamical}. Empowerment methods, on the other hand, develop skills that maximally cover the state space, often restricted to a few of its dimensions \shortcite<\eg the x-y space in navigation tasks>{achiam_variational_2018,eysenbach2018diversity,campos_explore_2020,sharma_dynamics-aware_2020}. 

These methods are limited to learn goal representations within a bounded, pre-defined space: the visual space, or the (sub-) state space. How to autonomously learn to represent the wild diversity of goals surveyed in Section~\ref{sec:survey_goal_rep} and discussed in Challenge \#1 remains an open question.

\subsection{Challenge \#3: Imagining Creative Goals}
Goal sampling methods surveyed in Section~\ref{sec:survey_generation} are all bound to sample goals \textit{within the distribution of known effects}. Indeed, the support of the goals distribution is either pre-defined  \shortcite<\eg >{schaul2015universal,andrychowicz2017hindsight,curious,li2019towards} or learned using a generative model \shortcite{goalgan,nair2018visual,nair2020contextual,pong2019skew} trained on previously experienced outcomes. On the other hand, humans can imagine creative goals beyond their past experience which, arguably, powers their exploration of the world. 

In this survey, one approach opened a path in this direction. The \imagine algorithm uses linguistic goal representation learned via social supervision and leverages the compositionality of language to imagine creative goals beyond its past experience \shortcite{imagine}. This is implemented by a simple mechanism detecting templates in known goals and recombining them to form new ones. This is in line with a recent line of work in developmental psychology arguing that human play might be about practicing to generate plans to solve imaginary problems  \shortcite{chu2020exploratory}.

Another way to achieve similar outcomes is to compose known goals with Boolean algebras, where new goals can be formed by composing existing atomic goals with negation, conjunction and disjunctions. The logical combinations of atomic goals was investigated in \shortciteA{tasse2020boolean,chitnis2020glib}, and \shortciteA{colas2020language,akakzia2020decstr}. The first approach represents the space of goals as a Boolean algebra, which allows immediate generalization to compositions of goals (\textsc{and}, \textsc{or}, \textsc{not}). The second approach considers using general symbolic and logic languages to express goals, but uses symbolic planning techniques that are not yet fully integrated in the goal-conditioned deep \rl framework. The third and fourth train a generative model of goals conditioned on language inputs. Because it generates discrete goals, it can compose language instructions by composing the finite sets of discrete goals associated to each instruction (\textsc{and} is the intersection, \textsc{or} the union etc). However, these works fall short of exploring the richness of goal compositionality and its various potential forms. \shortciteA{tasse2020boolean} seem to be limited to specific goals as target features, while \shortciteA{akakzia2020decstr} requires discrete goals. Finally, \shortciteA{barreto_option_2019} proposes to target new goals that are represented by linear combination of pseudo-rewards called \textit{cumulants}. They use the option framework and show that an agent that masters a set of options associated with cumulants can generalize to any new behavior induced by a linear combination of those known cumulants. 

\subsection{Challenge \#4: Composing Skills for Better Generalization}

Although this survey focuses on goal-related mechanisms, autotelic agents also need to learn to achieve their goals. Progress in this direction directly relies on progress in standard \rl and goal-conditioned \rl. In particular, autotelic agents would considerably benefit from better generalization and skill composition. Indeed, as the set of goals agents can target grows, it becomes more and more crucial that agents can efficiently transfer knowledge between skills, infer new skills from the ones they already master and compose skills to form more complex ones. Although hierarchical \rl approach learn to compose skills sequentially, concurrent skill composition remains under-explored.

\subsection{Challenge \#6: Leveraging Socio-Cultural Environments}
Decades of research in psychology, philosophy, linguistics and robotics have demonstrated the crucial importance of rich socio-cultural environments in human development \shortcite{vygotsky_thought_1934,whorf_language_1956,wood_role_1976,rumelhart_sequential_1986,berk_why_1994,clark_being_1998,tomasello_cultural_1999,tomasello_constructing_2009,zlatev_epigenesis_2001,carruthers_modularity_2002,dautenhahn_embodied_2002,lindblom_social_2003,mirolli_towards_2011,lupyan_what_2012}. However, modern \ai may have lost track of these insights. Deep reinforcement learning rarely considers social interactions and, when it does, models them as direct teaching; depriving agents of all autonomy. A recent discussion of this problem and an argument for the need of agents that are both autonomous and teachable can be found in a concurrent work \cite{sigaud2021towards}. As we embed autotelic agents in richer socio-cultural worlds and let them interact with humans, they might start to learn goal representations that are meaningful for us, in our society.

%\subsection{Language as a Tool for Creative Goal Generation}
%\label{sec:future_language_goals}
%Whether they are specific or abstract, time-specific or time-extended, whether they represent mixture of objectives, constraints, or logical combinations, all goals can be expressed easily by humans through language. Language, thus, seems like the ideal candidate to express goals in \rl agents. So far, language was only used to formulate a few forms of goals (see Section~\ref{sec:survey_goal_rep}). In the future, it might be used to express any type of goals. Recurrent Neural Networks (\rnn) \shortcite{elman}, Deep Sets \shortcite{deepset}, Graph Neural Networks (\gnn) or Transformers are all architectures that benefits from inductive biases and could be leveraged to facilitate new forms of goal representations (time-extended, set-based, relational etc.). As a learning signal, \shortciteA{imagine} propose to leverage descriptive sentences from social partners as a way to train goal representations. Mixing self-supervised learning and language inputs from humans as in \shortciteA{lynch2020grounding} might be the way forward. 

% % % % % % % % % % % % % % % % % % % % % % % % % % %
% Discussion \& Conclusion
% % % % % % % % % % % % % % % % % % % % % % % % % % % 
\section{Discussion \& Conclusion}

This paper defined the intrinsically motivated skills acquisition problem and proposed to view autotelic \rl algorithms or \rlimgep as computational tools to tackle it. These methods belong to the new field of \textit{developmental reinforcement learning}, the intersection of the developmental robotics and \rl fields. We reviewed current goal-conditioned \rl approaches under the lens of autotelic agents that learn to represent and generate their own goals in addition of learning to achieve them.

We propose a new general definition of the \textit{goal} construct: a pair of compact goal representation and an associated goal-achievement function. Interestingly, this viewpoint allowed us to categorize some \rl approaches as goal-conditioned, even though the original papers did not explicitly acknowledge it. For instance, we view the Never Give Up \shortcite{badia2020never} and Agent 57 \shortcite{badia2020agent57} architectures as goal-conditioned, because agents actively select parameters affecting the task at hand (parameter mixing extrinsic and intrinsic objectives, discount factor) and see their behavior affected by this choice (goal-conditioned policies).

This point of view also offers a direction for future research. Autotelic agents need to learn to represent goals and to measure goal achievement. Future research could extend the diversity of considered goal representations, investigate novel reward function architectures and inductive biases to allow time-extended goals, goal composition and to improve generalization.

%\todo{Maybe discuss the link with the GVF from Sutton and his question/answer stuff}\\
The general vision we convey in this paper builds on the metaphor of the learning agent as a curious scientist. A scientist that would formulate hypotheses about the world and explore it to find out whether they are true. A scientist that would ask questions, and setup intermediate goals to explore the world and find answers. A scientist that would set challenges to itself to learn about the world, to discover new ways to interact with it and to grow its collection of skills and knowledge. Such a scientist could decide of its own agenda. It would not need to be instructed and could be guided only by its curiosity, by its desire to discover new information and to master new skills. Autotelic agents should nonetheless be immersed in complex socio-cultural environment, just like humans are. In contact with humans, they could learn to represent goals that humans and society care about. 

\clearpage
%\begin{landscape}
%\end{landscape}
\renewcommand{\arraystretch}{1.3}
\begin{table*}[t!]
\tiny
\centering

\begin{tabular}{l|V{2cm}V{2cm}V{2cm}V{2cm}}
\hline
\multirow{2}{*}{\textbf{Approach}} & \multirow{2}{*}{\textbf{Goal Type}} & \multirow{2}{*}{\textbf{\shortstack{Goal \\ Rep.}}} & \multirow{2}{*}{\textbf{\shortstack{Reward\\ Function}}} & \multirow{2}{*}{\textbf{\shortstack{Goal sampling \\ strategy}}} \\
\\
\hline
\multicolumn{5}{l}{\textbf{RL-IMGEPs that assume goal embeddings and reward functions}}\\
\hline
\cite{fournier2018accuracy} & Target features (+tolerance) & Pre-def & Pre-def & \lp-Based\\
\textsc{hac} \cite{levy2018hierarchical} & Target features & Pre-def & Pre-def & \hrl \\
\textsc{hiro} \cite{nachum2018data} & Target features & Pre-def & Pre-def & \hrl \\
\textbf{CURIOUS} \cite{curious} & Target features & Pre-def & Pre-def & \lp-based\\
\textbf{CLIC} \cite{fournier2019clic} & Target features & Pre-def & Pre-def & \lp-based\\
\textbf{CWYC} \cite{blaes2019control} & Target features & Pre-def & Pre-def & \lp-based + surprise\\
\textsc{go-explore} \cite{ecoffet2020first} & Target features & Pre-def & Pre-def & Novelty \\
\textsc{ngu} \cite{badia2020never} & Objectives balance & Pre-def & Pre-def & Uniform \\
\textsc{agent} 57 \cite{badia2020agent57} & Objectives balance & Pre-def & Pre-def & Meta-learned \\ 
\textbf{DECSTR} \cite{akakzia2020decstr} & Binary problem & Pre-def & Pre-def & \lp-based \\
\textsc{slide} \cite{Fang-RSS-21} & Skill index & Pre-def & Pre-def & Novelty (PCG)\\
\textsc{XLand OEL} \cite{team2021open} & Binary problem & Pre-def & Pre-def & Intermediate difficulty \\
\hline
\multicolumn{5}{l}{\textbf{RL-IMGEPs that learn their goal embedding and assume reward functions}}\\
\hline
%\cite{sutton2011horde} & & & & \\
\textsc{rig} \cite{nair2018visual} & Target features (images) & Learned (\vae) & Pre-def & From \vae prior \\
\textsc{goalgan} \cite{goalgan} & Target features & Pre-def + GAN & Pre-def & Intermediate difficulty\\
\cite{florensa2019selfsupervised} & Target features (images) & Learned (\vae) & Pre-def & From \vae prior \\
\textsc{skew-fit} \cite{pong2019skew} & Target features (images) & Learned (\vae) & Pre-def & Diversity \\
\textsc{setter-solver} \cite{settersolver} & Target features (images) & Learned (Gen. model) & Pre-def & Uniform difficulty \\
\textsc{mega} \cite{pitis2020maximum} & Target features (images) & Learned (\vae) & Pre-def & Novelty \\
\textsc{cc-rig} \cite{nair2020contextual} & Target features (images) & Learned (\vae)  & Pre-def & From \vae prior \\
\textsc{amigo} \cite{campero2020learning} & Target features (images) & Learned (with policy) & Pre-def  & Adversarial \\ 
\textbf{GRIMGEP} \cite{kovac2020grimgep} & Target features (images) & Learned (with policy) & Pre-def & Diversity and ALP \\
\hline
\multicolumn{5}{l}{\textbf{Full RL-IMGEPs}}\\
\hline
\textsc{discern} \cite{warde2018unsupervised} & Target features (images) & Learned (with policy) & Learned (similarity) & Diversity \\
\textsc{diayn} \cite{eysenbach2018diversity} & Discrete skills & Learned (with policy) & Learned (discriminability) & Uniform \\
\cite{hartikainen2019dynamical} & Target features (images) & Learned (with policy) & Learned (distance) & Intermediate difficulty  \\
\cite{venkattaramanujam2019self} & Target features (images) & Learned (with policy) & Learned (distance) & Intermediate difficulty \\
\textbf{IMAGINE} \cite{imagine} & Binary problem (language) & Learned (with reward) & Learned & Uniform + Diversity \\
\textsc{vgcrl} \cite{choi_variational_2021} & Target features & Learned  & Learned & Empowerment \\
\end{tabular}
\caption{
\small \textbf{A classification of autotelic RL-IMGEP approaches.} \small Autotelic approaches require agents to sample their own goals. The proposed classification groups algorithms depending on their degree of autonomy: 1) \rlimgeps that rely on pre-defined goal representations (embeddings and reward functions); 2) \rlimgeps that rely on pre-defined reward functions but learn goal embeddings and 3) \rlimgeps that learn complete goal representations (embeddings and reward functions). For each algorithm, we report the type of goals being pursued (see Section~\ref{sec:survey_goal_rep}), whether goal embeddings are learned (Section~\ref{sec:survey_learning_goal_rep}), whether reward functions are learned (Section~\ref{sec:survey_learning_goal_rep_rew}) and how goals are sampled (Section~\ref{sec:survey_generation}). We mark in bold algorithms that use a developmental approaches and explicitly pursue the intrinsically motivated skills acquisition problem.}
\label{tab:bigtable}
\end{table*}
\clearpage

\vskip 0.2in
\bibliography{biblio}

\begin{thebibliography}{}

\bibitem[\protect\BCAY{Abramson, Ahuja, Brussee, Carnevale, Cassin, Clark,
  Dudzik, Georgiev, Guy, Harley, Hill, Hung, Kenton, Landon, Lillicrap,
  Mathewson, Muldal, Santoro, Savinov, Varma, Wayne, Wong, Yan,\ \BBA\
  Zhu}{Abramson et~al.}{2020}]{abramson_imitating_2020}
Abramson, J., Ahuja, A., Brussee, A., Carnevale, F., Cassin, M., Clark, S.,
  Dudzik, A., Georgiev, P., Guy, A., Harley, T., Hill, F., Hung, A., Kenton,
  Z., Landon, J., Lillicrap, T., Mathewson, K., Muldal, A., Santoro, A.,
  Savinov, N., Varma, V., Wayne, G., Wong, N., Yan, C., \BBA\ Zhu, R.
  \BBOP2020\BBCP.
\newblock \BBOQ Imitating {Interactive} {Intelligence}\BBCQ.

\bibitem[\protect\BCAY{Achiam, Edwards, Amodei,\ \BBA\ Abbeel}{Achiam
  et~al.}{2018}]{achiam_variational_2018}
Achiam, J., Edwards, H., Amodei, D., \BBA\ Abbeel, P. \BBOP2018\BBCP.
\newblock \BBOQ Variational option discovery algorithms\BBCQ\
\newblock ArXiv - {abs/1807.10299}.

\bibitem[\protect\BCAY{Achiam\ \BBA\ Sastry}{Achiam\ \BBA\
  Sastry}{2017}]{achiam2017surprise}
Achiam, J.\BBACOMMA\  \BBA\ Sastry, S. \BBOP2017\BBCP.
\newblock \BBOQ Surprise-based intrinsic motivation for deep reinforcement
  learning\BBCQ\
\newblock ArXiv - {abs/1703.01732}.

\bibitem[\protect\BCAY{Akakzia, Colas, Oudeyer, Chetouani,\ \BBA\
  Sigaud}{Akakzia et~al.}{2021}]{akakzia2020decstr}
Akakzia, A., Colas, C., Oudeyer, P.-Y., Chetouani, M., \BBA\ Sigaud, O.
  \BBOP2021\BBCP.
\newblock \BBOQ {DECSTR}: Learning goal-directed abstract behaviors using
  pre-verbal spatial predicates in intrinsically motivated agents\BBCQ\
\newblock In {\Bem Proc. of ICLR}.

\bibitem[\protect\BCAY{Andreas, Rohrbach, Darrell,\ \BBA\ Klein}{Andreas
  et~al.}{2016}]{Andreas_2016}
Andreas, J., Rohrbach, M., Darrell, T., \BBA\ Klein, D. \BBOP2016\BBCP.
\newblock \BBOQ Neural module networks\BBCQ\
\newblock In {\Bem 2016 {IEEE} Conference on Computer Vision and Pattern
  Recognition, {CVPR} 2016, Las Vegas, NV, USA, June 27-30, 2016}, \BPGS\
  39--48. {IEEE} Computer Society.

\bibitem[\protect\BCAY{Andrychowicz, Crow, Ray, Schneider, Fong, Welinder,
  McGrew, Tobin, Abbeel,\ \BBA\ Zaremba}{Andrychowicz
  et~al.}{2017}]{andrychowicz2017hindsight}
Andrychowicz, M., Crow, D., Ray, A., Schneider, J., Fong, R., Welinder, P.,
  McGrew, B., Tobin, J., Abbeel, P., \BBA\ Zaremba, W. \BBOP2017\BBCP.
\newblock \BBOQ Hindsight experience replay\BBCQ\
\newblock In {\Bem Proc. of NeurIPS}, \BPGS\ 5048--5058.

\bibitem[\protect\BCAY{Asada, Hosoda, Kuniyoshi, Ishiguro, Inui, Yoshikawa,
  Ogino,\ \BBA\ Yoshida}{Asada et~al.}{2009}]{asada2009cognitive}
Asada, M., Hosoda, K., Kuniyoshi, Y., Ishiguro, H., Inui, T., Yoshikawa, Y.,
  Ogino, M., \BBA\ Yoshida, C. \BBOP2009\BBCP.
\newblock \BBOQ Cognitive developmental robotics: A survey\BBCQ\
\newblock {\Bem IEEE transactions on autonomous mental development}, {\Bem
  1\/}(1), 12--34.

\bibitem[\protect\BCAY{Bacon, Harb,\ \BBA\ Precup}{Bacon
  et~al.}{2017}]{bacon2017option}
Bacon, P., Harb, J., \BBA\ Precup, D. \BBOP2017\BBCP.
\newblock \BBOQ The option-critic architecture\BBCQ\
\newblock In {\Bem Proc. of AAAI}, \BPGS\ 1726--1734.

\bibitem[\protect\BCAY{Badia, Piot, Kapturowski, Sprechmann, Vitvitskyi, Guo,\
  \BBA\ Blundell}{Badia et~al.}{2020a}]{badia2020agent57}
Badia, A.~P., Piot, B., Kapturowski, S., Sprechmann, P., Vitvitskyi, A., Guo,
  Z.~D., \BBA\ Blundell, C. \BBOP2020a\BBCP.
\newblock \BBOQ Agent57: Outperforming the atari human benchmark\BBCQ\
\newblock In {\Bem Proc. of ICML}, \lowercase{\BVOL}\ 119, \BPGS\ 507--517.

\bibitem[\protect\BCAY{Badia, Sprechmann, Vitvitskyi, Guo, Piot, Kapturowski,
  Tieleman, Arjovsky, Pritzel, Bolt,\ \BBA\ Blundell}{Badia
  et~al.}{2020b}]{badia2020never}
Badia, A.~P., Sprechmann, P., Vitvitskyi, A., Guo, D., Piot, B., Kapturowski,
  S., Tieleman, O., Arjovsky, M., Pritzel, A., Bolt, A., \BBA\ Blundell, C.
  \BBOP2020b\BBCP.
\newblock \BBOQ Never give up: Learning directed exploration strategies\BBCQ\
\newblock In {\Bem Proc. of ICLR}.

\bibitem[\protect\BCAY{Bahdanau, Hill, Leike, Hughes, Hosseini, Kohli,\ \BBA\
  Grefenstette}{Bahdanau et~al.}{2019a}]{bahdanau2018learning}
Bahdanau, D., Hill, F., Leike, J., Hughes, E., Hosseini, S.~A., Kohli, P.,
  \BBA\ Grefenstette, E. \BBOP2019a\BBCP.
\newblock \BBOQ Learning to understand goal specifications by modelling
  reward\BBCQ\
\newblock In {\Bem Proc. of ICLR}.

\bibitem[\protect\BCAY{Bahdanau, Murty, Noukhovitch, Nguyen, de~Vries,\ \BBA\
  Courville}{Bahdanau et~al.}{2019b}]{bahdanau2018systematic}
Bahdanau, D., Murty, S., Noukhovitch, M., Nguyen, T.~H., de~Vries, H., \BBA\
  Courville, A.~C. \BBOP2019b\BBCP.
\newblock \BBOQ Systematic generalization: What is required and can it be
  learned?\BBCQ\
\newblock In {\Bem Proc. of ICLR}.

\bibitem[\protect\BCAY{Baranes\ \BBA\ Oudeyer}{Baranes\ \BBA\
  Oudeyer}{2009a}]{baranes2009proximo}
Baranes, A.\BBACOMMA\  \BBA\ Oudeyer, P.-Y. \BBOP2009a\BBCP.
\newblock \BBOQ Proximo-distal competence based curiosity-driven
  exploration\BBCQ\
\newblock In {\Bem Learning, in International Conference on Epigenetic
  Robotics, Italie. Citeseer}. Citeseer.

\bibitem[\protect\BCAY{Baranes\ \BBA\ Oudeyer}{Baranes\ \BBA\
  Oudeyer}{2009b}]{baranes2009r}
Baranes, A.\BBACOMMA\  \BBA\ Oudeyer, P.-Y. \BBOP2009b\BBCP.
\newblock \BBOQ R-iac: Robust intrinsically motivated exploration and active
  learning\BBCQ\
\newblock In {\Bem IEEE Transactions on Autonomous Mental Development},
  \lowercase{\BVOL}~1, \BPGS\ 155--169. IEEE.

\bibitem[\protect\BCAY{Baranes\ \BBA\ Oudeyer}{Baranes\ \BBA\
  Oudeyer}{2010}]{baranes2010intrinsically}
Baranes, A.\BBACOMMA\  \BBA\ Oudeyer, P.-Y. \BBOP2010\BBCP.
\newblock \BBOQ Intrinsically motivated goal exploration for active motor
  learning in robots: A case study\BBCQ\
\newblock In {\Bem 2010 IEEE/RSJ International Conference on Intelligent Robots
  and Systems}, \BPGS\ 1766--1773. IEEE.

\bibitem[\protect\BCAY{Baranes\ \BBA\ Oudeyer}{Baranes\ \BBA\
  Oudeyer}{2013}]{baranes2013active}
Baranes, A.\BBACOMMA\  \BBA\ Oudeyer, P.-Y. \BBOP2013\BBCP.
\newblock \BBOQ Active learning of inverse models with intrinsically motivated
  goal exploration in robots\BBCQ\
\newblock {\Bem Robotics and Autonomous Systems}, {\Bem 61\/}(1), 49--73.

\bibitem[\protect\BCAY{Barreto, Borsa, Hou, Comanici, Aygün, Hamel, Toyama,
  hunt, Mourad, Silver,\ \BBA\ Precup}{Barreto
  et~al.}{2019}]{barreto_option_2019}
Barreto, A., Borsa, D., Hou, S., Comanici, G., Aygün, E., Hamel, P., Toyama,
  D., hunt, J., Mourad, S., Silver, D., \BBA\ Precup, D. \BBOP2019\BBCP.
\newblock \BBOQ The option keyboard: Combining skills in reinforcement
  learning\BBCQ\
\newblock In {\Bem Proc. of NeurIPS}, \lowercase{\BVOL}~32.

\bibitem[\protect\BCAY{Barreto, Dabney, Munos, Hunt, Schaul, Silver,\ \BBA\ van
  Hasselt}{Barreto et~al.}{2017}]{NIPS2017_350db081}
Barreto, A., Dabney, W., Munos, R., Hunt, J.~J., Schaul, T., Silver, D., \BBA\
  van Hasselt, H. \BBOP2017\BBCP.
\newblock \BBOQ Successor features for transfer in reinforcement learning\BBCQ\
\newblock In {\Bem Proc. of NeurIPS}, \BPGS\ 4055--4065.

\bibitem[\protect\BCAY{Barreto, Hou, Borsa, Silver,\ \BBA\ Precup}{Barreto
  et~al.}{2020}]{barreto2020fast}
Barreto, A., Hou, S., Borsa, D., Silver, D., \BBA\ Precup, D. \BBOP2020\BBCP.
\newblock \BBOQ Fast reinforcement learning with generalized policy
  updates\BBCQ\
\newblock {\Bem Proceedings of the National Academy of Sciences}, {\Bem
  117\/}(48), 30079--30087.

\bibitem[\protect\BCAY{Bellemare, Candido, Castro, Gong, Machado, Moitra,
  Ponda,\ \BBA\ Wang}{Bellemare et~al.}{2020}]{bellemare2020autonomous}
Bellemare, M.~G., Candido, S., Castro, P.~S., Gong, J., Machado, M.~C., Moitra,
  S., Ponda, S.~S., \BBA\ Wang, Z. \BBOP2020\BBCP.
\newblock \BBOQ Autonomous navigation of stratospheric balloons using
  reinforcement learning\BBCQ\
\newblock {\Bem Nature}, {\Bem 588\/}(7836), 77--82.

\bibitem[\protect\BCAY{Bellemare, Srinivasan, Ostrovski, Schaul, Saxton,\ \BBA\
  Munos}{Bellemare et~al.}{2016}]{bellemare2016unifying}
Bellemare, M.~G., Srinivasan, S., Ostrovski, G., Schaul, T., Saxton, D., \BBA\
  Munos, R. \BBOP2016\BBCP.
\newblock \BBOQ Unifying count-based exploration and intrinsic motivation\BBCQ\
\newblock In {\Bem Proc. of NeurIPS}, \BPGS\ 1471--1479.

\bibitem[\protect\BCAY{Berk}{Berk}{1994}]{berk_why_1994}
Berk, L.~E. \BBOP1994\BBCP.
\newblock \BBOQ Why {Children} {Talk} to {Themselves}\BBCQ\
\newblock {\Bem Scientific American}, {\Bem 271\/}(5), 78--83.

\bibitem[\protect\BCAY{Berlyne}{Berlyne}{1966}]{berlyne1966curiosity}
Berlyne, D.~E. \BBOP1966\BBCP.
\newblock \BBOQ Curiosity and exploration\BBCQ\
\newblock {\Bem Science}, {\Bem 153\/}(3731), 25--33.

\bibitem[\protect\BCAY{Berseth, Geng, Devin, Finn, Jayaraman,\ \BBA\
  Levine}{Berseth et~al.}{2019}]{berseth2019smirl}
Berseth, G., Geng, D., Devin, C., Finn, C., Jayaraman, D., \BBA\ Levine, S.
  \BBOP2019\BBCP.
\newblock \BBOQ Smirl: Surprise minimizing rl in dynamic environments\BBCQ\
\newblock ArXiv - abs/1912.05510.

\bibitem[\protect\BCAY{Blaes, Pogancic, Zhu,\ \BBA\ Martius}{Blaes
  et~al.}{2019}]{blaes2019control}
Blaes, S., Pogancic, M.~V., Zhu, J., \BBA\ Martius, G. \BBOP2019\BBCP.
\newblock \BBOQ Control what you can: Intrinsically motivated task-planning
  agent\BBCQ\
\newblock In {\Bem Proc. of NeurIPS}, \BPGS\ 12520--12531.

\bibitem[\protect\BCAY{Brown, Mann, Ryder, Subbiah, Kaplan, Dhariwal,
  Neelakantan, Shyam, Sastry, Askell, Agarwal, Herbert{-}Voss, Krueger,
  Henighan, Child, Ramesh, Ziegler, Wu, Winter, Hesse, Chen, Sigler, Litwin,
  Gray, Chess, Clark, Berner, McCandlish, Radford, Sutskever,\ \BBA\
  Amodei}{Brown et~al.}{2020}]{brown2020language}
Brown, T.~B., Mann, B., Ryder, N., Subbiah, M., Kaplan, J., Dhariwal, P.,
  Neelakantan, A., Shyam, P., Sastry, G., Askell, A., Agarwal, S.,
  Herbert{-}Voss, A., Krueger, G., Henighan, T., Child, R., Ramesh, A.,
  Ziegler, D.~M., Wu, J., Winter, C., Hesse, C., Chen, M., Sigler, E., Litwin,
  M., Gray, S., Chess, B., Clark, J., Berner, C., McCandlish, S., Radford, A.,
  Sutskever, I., \BBA\ Amodei, D. \BBOP2020\BBCP.
\newblock \BBOQ Language models are few-shot learners\BBCQ\
\newblock In {\Bem Proc. of NeurIPS}.

\bibitem[\protect\BCAY{Burda, Edwards, Storkey,\ \BBA\ Klimov}{Burda
  et~al.}{2019}]{burda2018exploration}
Burda, Y., Edwards, H., Storkey, A.~J., \BBA\ Klimov, O. \BBOP2019\BBCP.
\newblock \BBOQ Exploration by random network distillation\BBCQ\
\newblock In {\Bem Proc. of ICLR}.

\bibitem[\protect\BCAY{Campero, Raileanu, K{\"{u}}ttler, Tenenbaum,
  Rockt{\"{a}}schel,\ \BBA\ Grefenstette}{Campero
  et~al.}{2021}]{campero2020learning}
Campero, A., Raileanu, R., K{\"{u}}ttler, H., Tenenbaum, J.~B.,
  Rockt{\"{a}}schel, T., \BBA\ Grefenstette, E. \BBOP2021\BBCP.
\newblock \BBOQ Learning with amigo: Adversarially motivated intrinsic
  goals\BBCQ\
\newblock In {\Bem Proc. of ICLR}.

\bibitem[\protect\BCAY{Campos, Trott, Xiong, Socher, Gir{\'{o}}{-}i{-}Nieto,\
  \BBA\ Torres}{Campos et~al.}{2020}]{campos_explore_2020}
Campos, V., Trott, A., Xiong, C., Socher, R., Gir{\'{o}}{-}i{-}Nieto, X., \BBA\
  Torres, J. \BBOP2020\BBCP.
\newblock \BBOQ Explore, discover and learn: Unsupervised discovery of
  state-covering skills\BBCQ\
\newblock In {\Bem Proc. of ICML}, \lowercase{\BVOL}\ 119, \BPGS\ 1317--1327.

\bibitem[\protect\BCAY{Cangelosi\ \BBA\ Schlesinger}{Cangelosi\ \BBA\
  Schlesinger}{2015}]{cangelosi2015developmental}
Cangelosi, A.\BBACOMMA\  \BBA\ Schlesinger, M. \BBOP2015\BBCP.
\newblock {\Bem Developmental Robotics: From Babies to Robots}.
\newblock MIT press.

\bibitem[\protect\BCAY{Carruthers}{Carruthers}{2002}]{carruthers_modularity_2002}
Carruthers, P. \BBOP2002\BBCP.
\newblock \BBOQ Modularity, {Language}, and the {Flexibility} of
  {Thought}\BBCQ\
\newblock {\Bem Behavioral and Brain Sciences}, {\Bem 25\/}(6), 705--719.

\bibitem[\protect\BCAY{Caruana}{Caruana}{1997}]{caruana1997multitask}
Caruana, R. \BBOP1997\BBCP.
\newblock \BBOQ Multitask learning\BBCQ\
\newblock {\Bem Machine learning}, {\Bem 28\/}(1), 41--75.

\bibitem[\protect\BCAY{Celik, Zhou, Li, Becker,\ \BBA\ Neumann}{Celik
  et~al.}{2021}]{celik_specializing_2021}
Celik, O., Zhou, D., Li, G., Becker, P., \BBA\ Neumann, G. \BBOP2021\BBCP.
\newblock \BBOQ Specializing {Versatile} {Skill} {Libraries} using {Local}
  {Mixture} of {Experts}\BBCQ\
\newblock In {\Bem 5th {Annual} {Conference} on {Robot} {Learning}}.

\bibitem[\protect\BCAY{Chan, Wu, Kiros, Fidler,\ \BBA\ Ba}{Chan
  et~al.}{2019}]{chan2019actrce}
Chan, H., Wu, Y., Kiros, J., Fidler, S., \BBA\ Ba, J. \BBOP2019\BBCP.
\newblock \BBOQ Actrce: Augmenting experience via teacher's advice for
  multi-goal reinforcement learning\BBCQ\
\newblock ArXiv - abs/1902.04546.

\bibitem[\protect\BCAY{Chaplot, Sathyendra, Pasumarthi, Rajagopal,\ \BBA\
  Salakhutdinov}{Chaplot et~al.}{2018}]{chaplot2017gatedattention}
Chaplot, D.~S., Sathyendra, K.~M., Pasumarthi, R.~K., Rajagopal, D., \BBA\
  Salakhutdinov, R. \BBOP2018\BBCP.
\newblock \BBOQ Gated-attention architectures for task-oriented language
  grounding\BBCQ\
\newblock In {\Bem Proc. of AAAI}, \BPGS\ 2819--2826.

\bibitem[\protect\BCAY{Charlesworth\ \BBA\ Montana}{Charlesworth\ \BBA\
  Montana}{2020}]{charlesworth2020plangan}
Charlesworth, H.\BBACOMMA\  \BBA\ Montana, G. \BBOP2020\BBCP.
\newblock \BBOQ Plangan: Model-based planning with sparse rewards and multiple
  goals\BBCQ\
\newblock In {\Bem Proc. of NeurIPS}.

\bibitem[\protect\BCAY{Chevalier-Boisvert, Bahdanau, Lahlou, Willems, Saharia,
  Nguyen,\ \BBA\ Bengio}{Chevalier-Boisvert
  et~al.}{2019}]{chevalier-boisvert2018babyai}
Chevalier-Boisvert, M., Bahdanau, D., Lahlou, S., Willems, L., Saharia, C.,
  Nguyen, T.~H., \BBA\ Bengio, Y. \BBOP2019\BBCP.
\newblock \BBOQ {Baby{AI}: First Steps Towards Grounded Language Learning With
  a Human In the Loop}\BBCQ\
\newblock In {\Bem International Conference on Learning Representations}.

\bibitem[\protect\BCAY{Chitnis, Silver, Tenenbaum, Kaelbling,\ \BBA\
  Lozano-P{\'e}rez}{Chitnis et~al.}{2021}]{chitnis2020glib}
Chitnis, R., Silver, T., Tenenbaum, J., Kaelbling, L.~P., \BBA\
  Lozano-P{\'e}rez, T. \BBOP2021\BBCP.
\newblock \BBOQ Glib: Efficient exploration for relational model-based
  reinforcement learning via goal-literal babbling\BBCQ\
\newblock In {\Bem AAAI}.

\bibitem[\protect\BCAY{Choi, Sharma, Lee, Levine,\ \BBA\ Gu}{Choi
  et~al.}{2021}]{choi_variational_2021}
Choi, J., Sharma, A., Lee, H., Levine, S., \BBA\ Gu, S.~S. \BBOP2021\BBCP.
\newblock \BBOQ Variational {Empowerment} as {Representation} {Learning} for
  {Goal}-{Based} {Reinforcement} {Learning}\BBCQ\
\newblock ArXiv - abs/2106.01404.

\bibitem[\protect\BCAY{Chu\ \BBA\ Schulz}{Chu\ \BBA\
  Schulz}{2020}]{chu2020exploratory}
Chu, J.\BBACOMMA\  \BBA\ Schulz, L. \BBOP2020\BBCP.
\newblock \BBOQ Exploratory play, rational action, and efficient search\BBCQ\
\newblock PsyArXiv.

\bibitem[\protect\BCAY{Chua, Calandra, McAllister,\ \BBA\ Levine}{Chua
  et~al.}{2018}]{chua2018deep}
Chua, K., Calandra, R., McAllister, R., \BBA\ Levine, S. \BBOP2018\BBCP.
\newblock \BBOQ Deep reinforcement learning in a handful of trials using
  probabilistic dynamics models\BBCQ\
\newblock In {\Bem Proc. of NeurIPS}, \BPGS\ 4759--4770.

\bibitem[\protect\BCAY{Cideron, Seurin, Strub,\ \BBA\ Pietquin}{Cideron
  et~al.}{2020}]{ther}
Cideron, G., Seurin, M., Strub, F., \BBA\ Pietquin, O. \BBOP2020\BBCP.
\newblock \BBOQ Higher: Improving instruction following with hindsight
  generation for experience replay\BBCQ\
\newblock In {\Bem 2020 IEEE Symposium Series on Computational Intelligence
  (SSCI)}, \BPGS\ 225--232. IEEE.

\bibitem[\protect\BCAY{Clark}{Clark}{1998}]{clark_being_1998}
Clark, A. \BBOP1998\BBCP.
\newblock {\Bem Being {There}: {Putting} {Brain}, {Body}, and {World}
  {Together} {Again}}.
\newblock MIT press.

\bibitem[\protect\BCAY{Codevilla, M\"{u}ller, L{\'o}pez, Koltun,\ \BBA\
  Dosovitskiy}{Codevilla et~al.}{2018}]{codevilla2018end}
Codevilla, F., M\"{u}ller, M., L{\'o}pez, A., Koltun, V., \BBA\ Dosovitskiy, A.
  \BBOP2018\BBCP.
\newblock \BBOQ End-to-end driving via conditional imitation learning\BBCQ\
\newblock In {\Bem 2018 IEEE International Conference on Robotics and
  Automation (ICRA)}, \BPGS\ 1--9. IEEE.

\bibitem[\protect\BCAY{Colas, Akakzia, Oudeyer, Chetouani,\ \BBA\ Sigaud}{Colas
  et~al.}{2020}]{colas2020language}
Colas, C., Akakzia, A., Oudeyer, P.-Y., Chetouani, M., \BBA\ Sigaud, O.
  \BBOP2020\BBCP.
\newblock \BBOQ Language-conditioned goal generation: a new approach to
  language grounding for rl\BBCQ\
\newblock ArXiv - abs/2006.07043.

\bibitem[\protect\BCAY{Colas, Hejblum, Rouillon, Thi{\'e}baut, Oudeyer,
  Moulin-Frier,\ \BBA\ Prague}{Colas et~al.}{2021}]{colas2020epidemioptim}
Colas, C., Hejblum, B., Rouillon, S., Thi{\'e}baut, R., Oudeyer, P.-Y.,
  Moulin-Frier, C., \BBA\ Prague, M. \BBOP2021\BBCP.
\newblock \BBOQ Epidemioptim: A toolbox for the optimization of control
  policies in epidemiological models\BBCQ\
\newblock {\Bem Journal of Artificial Intelligence Research}, {\Bem 71}.

\bibitem[\protect\BCAY{Colas, Karch, Lair, Dussoux, Moulin{-}Frier, Dominey,\
  \BBA\ Oudeyer}{Colas et~al.}{2020a}]{imagine}
Colas, C., Karch, T., Lair, N., Dussoux, J., Moulin{-}Frier, C., Dominey,
  P.~F., \BBA\ Oudeyer, P. \BBOP2020a\BBCP.
\newblock \BBOQ Language as a cognitive tool to imagine goals in curiosity
  driven exploration\BBCQ\
\newblock In {\Bem Proc. of NeurIPS}.

\bibitem[\protect\BCAY{Colas, Madhavan, Huizinga,\ \BBA\ Clune}{Colas
  et~al.}{2020b}]{colas2020scaling}
Colas, C., Madhavan, V., Huizinga, J., \BBA\ Clune, J. \BBOP2020b\BBCP.
\newblock \BBOQ Scaling map-elites to deep neuroevolution\BBCQ\
\newblock In {\Bem Proc. of GECCO}, \BPGS\ 67--75.

\bibitem[\protect\BCAY{Colas, Oudeyer, Sigaud, Fournier,\ \BBA\
  Chetouani}{Colas et~al.}{2019}]{curious}
Colas, C., Oudeyer, P., Sigaud, O., Fournier, P., \BBA\ Chetouani, M.
  \BBOP2019\BBCP.
\newblock \BBOQ {CURIOUS:} intrinsically motivated modular multi-goal
  reinforcement learning\BBCQ\
\newblock In {\Bem Proc. of ICML}, \lowercase{\BVOL}~97, \BPGS\ 1331--1340.

\bibitem[\protect\BCAY{Colas, Sigaud,\ \BBA\ Oudeyer}{Colas
  et~al.}{2018}]{geppg}
Colas, C., Sigaud, O., \BBA\ Oudeyer, P. \BBOP2018\BBCP.
\newblock \BBOQ {GEP-PG:} decoupling exploration and exploitation in deep
  reinforcement learning algorithms\BBCQ\
\newblock In {\Bem Proc. of ICML}, \lowercase{\BVOL}~80, \BPGS\ 1038--1047.

\bibitem[\protect\BCAY{Dai, Xu, Hofmann,\ \BBA\ Williams}{Dai
  et~al.}{2020}]{dai_empowerment-based_2020}
Dai, S., Xu, W., Hofmann, A., \BBA\ Williams, B. \BBOP2020\BBCP.
\newblock \BBOQ An {Empowerment}-based {Solution} to {Robotic} {Manipulation}
  {Tasks} with {Sparse} {Rewards}\BBCQ\
\newblock ArXiv - abs/2010.07986.

\bibitem[\protect\BCAY{Das, Datta, Gkioxari, Lee, Parikh,\ \BBA\ Batra}{Das
  et~al.}{2018}]{das2018embodied}
Das, A., Datta, S., Gkioxari, G., Lee, S., Parikh, D., \BBA\ Batra, D.
  \BBOP2018\BBCP.
\newblock \BBOQ Embodied question answering\BBCQ\
\newblock In {\Bem 2018 {IEEE} Conference on Computer Vision and Pattern
  Recognition, {CVPR} 2018, Salt Lake City, UT, USA, June 18-22, 2018}, \BPGS\
  1--10. {IEEE} Computer Society.

\bibitem[\protect\BCAY{Dautenhahn, Ogden,\ \BBA\ Quick}{Dautenhahn
  et~al.}{2002}]{dautenhahn_embodied_2002}
Dautenhahn, K., Ogden, B., \BBA\ Quick, T. \BBOP2002\BBCP.
\newblock \BBOQ From {Embodied} to {Socially} {Embedded} {Agents} –
  {Implications} for {Interaction}-{Aware} {Robots}\BBCQ\
\newblock {\Bem Cognitive Systems Research}, {\Bem 3\/}(3), 397--428.

\bibitem[\protect\BCAY{Dayan\ \BBA\ Hinton}{Dayan\ \BBA\
  Hinton}{1993}]{dayan1993feudal}
Dayan, P.\BBACOMMA\  \BBA\ Hinton, G.~E. \BBOP1993\BBCP.
\newblock \BBOQ Feudal reinforcement learning\BBCQ\
\newblock In {\Bem Advances in neural information processing systems}, \BPGS\
  271--278.

\bibitem[\protect\BCAY{Dayan, Hinton, Neal,\ \BBA\ Zemel}{Dayan
  et~al.}{1995}]{dayan_helmholtz_1995}
Dayan, P., Hinton, G.~E., Neal, R.~M., \BBA\ Zemel, R.~S. \BBOP1995\BBCP.
\newblock \BBOQ The {Helmholtz} {Machine}\BBCQ\
\newblock {\Bem Neural Computation}, {\Bem 7\/}(5), 889--904.

\bibitem[\protect\BCAY{Devlin, Chang, Lee,\ \BBA\ Toutanova}{Devlin
  et~al.}{2019}]{devlin2019bert}
Devlin, J., Chang, M.-W., Lee, K., \BBA\ Toutanova, K. \BBOP2019\BBCP.
\newblock \BBOQ {BERT}: Pre-training of deep bidirectional transformers for
  language understanding\BBCQ\
\newblock In {\Bem Proc. of NAACL-HLT}, \BPGS\ 4171--4186. Association for
  Computational Linguistics.

\bibitem[\protect\BCAY{Ding, Florensa, Abbeel,\ \BBA\ Phielipp}{Ding
  et~al.}{2019}]{ding_imitation_2019}
Ding, Y., Florensa, C., Abbeel, P., \BBA\ Phielipp, M. \BBOP2019\BBCP.
\newblock \BBOQ Goal-conditioned imitation learning\BBCQ\
\newblock In {\Bem Proc. of NeurIPS}, \BPGS\ 15298--15309.

\bibitem[\protect\BCAY{Ecoffet, Huizinga, Lehman, Stanley,\ \BBA\
  Clune}{Ecoffet et~al.}{2021}]{ecoffet2020first}
Ecoffet, A., Huizinga, J., Lehman, J., Stanley, K.~O., \BBA\ Clune, J.
  \BBOP2021\BBCP.
\newblock \BBOQ First return, then explore\BBCQ\
\newblock {\Bem Nature}, {\Bem 590\/}(7847), 580--586.

\bibitem[\protect\BCAY{Elliot\ \BBA\ Fryer}{Elliot\ \BBA\
  Fryer}{2008}]{elliot2008goal}
Elliot, A.~J.\BBACOMMA\  \BBA\ Fryer, J.~W. \BBOP2008\BBCP.
\newblock \BBOQ The goal construct in psychology\BBCQ\
\newblock {\Bem Handbook of motivation science}, {\Bem 18}, 235--250.

\bibitem[\protect\BCAY{Elman}{Elman}{1993}]{elman}
Elman, J.~L. \BBOP1993\BBCP.
\newblock \BBOQ Learning and development in neural networks: the importance of
  starting small\BBCQ\
\newblock {\Bem Cognition}, {\Bem 48\/}(1), 71 -- 99.

\bibitem[\protect\BCAY{Etcheverry, Moulin{-}Frier,\ \BBA\ Oudeyer}{Etcheverry
  et~al.}{2020}]{etcheverry_hierarchically-organized_2020}
Etcheverry, M., Moulin{-}Frier, C., \BBA\ Oudeyer, P. \BBOP2020\BBCP.
\newblock \BBOQ Hierarchically organized latent modules for exploratory search
  in morphogenetic systems\BBCQ\
\newblock In {\Bem Proc. of NeurIPS}.

\bibitem[\protect\BCAY{Eysenbach, Geng, Levine,\ \BBA\ Salakhutdinov}{Eysenbach
  et~al.}{2020}]{eysenbach2020rewriting}
Eysenbach, B., Geng, X., Levine, S., \BBA\ Salakhutdinov, R.~R. \BBOP2020\BBCP.
\newblock \BBOQ Rewriting history with inverse {RL:} hindsight inference for
  policy improvement\BBCQ\
\newblock In {\Bem Proc. of NeurIPS}.

\bibitem[\protect\BCAY{Eysenbach, Gupta, Ibarz,\ \BBA\ Levine}{Eysenbach
  et~al.}{2019}]{eysenbach2018diversity}
Eysenbach, B., Gupta, A., Ibarz, J., \BBA\ Levine, S. \BBOP2019\BBCP.
\newblock \BBOQ Diversity is all you need: Learning skills without a reward
  function\BBCQ\
\newblock In {\Bem Proc. of ICLR}.

\bibitem[\protect\BCAY{Fang, Zhu, Savarese,\ \BBA\ Fei-Fei}{Fang
  et~al.}{2021}]{Fang-RSS-21}
Fang, K., Zhu, Y., Savarese, S., \BBA\ Fei-Fei, L. \BBOP2021\BBCP.
\newblock \BBOQ {Discovering Generalizable Skills via Automated Generation of
  Diverse Tasks}\BBCQ\
\newblock In {\Bem Proceedings of Robotics: Science and Systems}.

\bibitem[\protect\BCAY{Florensa, Degrave, Heess, Springenberg,\ \BBA\
  Riedmiller}{Florensa et~al.}{2019}]{florensa2019selfsupervised}
Florensa, C., Degrave, J., Heess, N., Springenberg, J.~T., \BBA\ Riedmiller, M.
  \BBOP2019\BBCP.
\newblock \BBOQ Self-supervised learning of image embedding for continuous
  control\BBCQ\
\newblock ArXiv - abs/1901.00943.

\bibitem[\protect\BCAY{Florensa, Held, Geng,\ \BBA\ Abbeel}{Florensa
  et~al.}{2018}]{goalgan}
Florensa, C., Held, D., Geng, X., \BBA\ Abbeel, P. \BBOP2018\BBCP.
\newblock \BBOQ Automatic goal generation for reinforcement learning
  agents\BBCQ\
\newblock In {\Bem Proc. of ICML}, \lowercase{\BVOL}~80, \BPGS\ 1514--1523.

\bibitem[\protect\BCAY{Forestier\ \BBA\ Oudeyer}{Forestier\ \BBA\
  Oudeyer}{2016}]{forestier2016modular}
Forestier, S.\BBACOMMA\  \BBA\ Oudeyer, P.-Y. \BBOP2016\BBCP.
\newblock \BBOQ Modular active curiosity-driven discovery of tool use\BBCQ\
\newblock In {\Bem Intelligent Robots and Systems (IROS), 2016 IEEE/RSJ
  International Conference on}, \BPGS\ 3965--3972. IEEE.

\bibitem[\protect\BCAY{Forestier, Portelas, Mollard,\ \BBA\ Oudeyer}{Forestier
  et~al.}{2017}]{imgep}
Forestier, S., Portelas, R., Mollard, Y., \BBA\ Oudeyer, P.-Y. \BBOP2017\BBCP.
\newblock \BBOQ Intrinsically motivated goal exploration processes with
  automatic curriculum learning\BBCQ\
\newblock ArXiv - abs/1708.02190.

\bibitem[\protect\BCAY{Fournier, Colas, Chetouani,\ \BBA\ Sigaud}{Fournier
  et~al.}{2021}]{fournier2019clic}
Fournier, P., Colas, C., Chetouani, M., \BBA\ Sigaud, O. \BBOP2021\BBCP.
\newblock \BBOQ Clic: Curriculum learning and imitation for object control in
  nonrewarding environments\BBCQ\
\newblock {\Bem IEEE Transactions on Cognitive and Developmental Systems},
  {\Bem 13\/}(2), 239--248.

\bibitem[\protect\BCAY{Fournier, Sigaud, Chetouani,\ \BBA\ Oudeyer}{Fournier
  et~al.}{2018}]{fournier2018accuracy}
Fournier, P., Sigaud, O., Chetouani, M., \BBA\ Oudeyer, P.-Y. \BBOP2018\BBCP.
\newblock \BBOQ Accuracy-based curriculum learning in deep reinforcement
  learning\BBCQ\
\newblock ArXiv - abs/1806.09614.

\bibitem[\protect\BCAY{Frans, Ho, Chen, Abbeel,\ \BBA\ Schulman}{Frans
  et~al.}{2018}]{frans2017meta}
Frans, K., Ho, J., Chen, X., Abbeel, P., \BBA\ Schulman, J. \BBOP2018\BBCP.
\newblock \BBOQ Meta learning shared hierarchies\BBCQ\
\newblock In {\Bem Proc. of ICLR}.

\bibitem[\protect\BCAY{Goodfellow, Pouget{-}Abadie, Mirza, Xu, Warde{-}Farley,
  Ozair, Courville,\ \BBA\ Bengio}{Goodfellow
  et~al.}{2014}]{goodfellow2014generative}
Goodfellow, I.~J., Pouget{-}Abadie, J., Mirza, M., Xu, B., Warde{-}Farley, D.,
  Ozair, S., Courville, A.~C., \BBA\ Bengio, Y. \BBOP2014\BBCP.
\newblock \BBOQ Generative adversarial nets\BBCQ\
\newblock In {\Bem Proc. of NeurIPS}, \BPGS\ 2672--2680.

\bibitem[\protect\BCAY{Gopnik, Meltzoff,\ \BBA\ Kuhl}{Gopnik
  et~al.}{1999}]{gopnik1999scientist}
Gopnik, A., Meltzoff, A.~N., \BBA\ Kuhl, P.~K. \BBOP1999\BBCP.
\newblock {\Bem The scientist in the crib: Minds, brains, and how children
  learn.}
\newblock William Morrow \& Co.

\bibitem[\protect\BCAY{Gottlieb\ \BBA\ Oudeyer}{Gottlieb\ \BBA\
  Oudeyer}{2018}]{gottlieb2018towards}
Gottlieb, J.\BBACOMMA\  \BBA\ Oudeyer, P.-Y. \BBOP2018\BBCP.
\newblock \BBOQ Towards a neuroscience of active sampling and curiosity\BBCQ\
\newblock {\Bem Nature Reviews Neuroscience}, {\Bem 19\/}(12), 758--770.

\bibitem[\protect\BCAY{Goyal, Lamb, Hoffmann, Sodhani, Levine, Bengio,\ \BBA\
  Sch{\"{o}}lkopf}{Goyal et~al.}{2021}]{goyal2019recurrent}
Goyal, A., Lamb, A., Hoffmann, J., Sodhani, S., Levine, S., Bengio, Y., \BBA\
  Sch{\"{o}}lkopf, B. \BBOP2021\BBCP.
\newblock \BBOQ Recurrent independent mechanisms\BBCQ\
\newblock In {\Bem Proc. of ICLR}.

\bibitem[\protect\BCAY{Gregor, Rezende,\ \BBA\ Wierstra}{Gregor
  et~al.}{2016}]{gregor2016variational}
Gregor, K., Rezende, D.~J., \BBA\ Wierstra, D. \BBOP2016\BBCP.
\newblock \BBOQ Variational intrinsic control\BBCQ\
\newblock ArXiv - abs/1611.07507.

\bibitem[\protect\BCAY{Hamrick, Friesen, Behbahani, Guez, Viola, Witherspoon,
  Anthony, Buesing, Velickovic,\ \BBA\ Weber}{Hamrick
  et~al.}{2021}]{hamrick_role_2020}
Hamrick, J.~B., Friesen, A.~L., Behbahani, F., Guez, A., Viola, F.,
  Witherspoon, S., Anthony, T., Buesing, L.~H., Velickovic, P., \BBA\ Weber, T.
  \BBOP2021\BBCP.
\newblock \BBOQ On the role of planning in model-based deep reinforcement
  learning\BBCQ\
\newblock In {\Bem Proc. of ICLR}.

\bibitem[\protect\BCAY{Hartikainen, Geng, Haarnoja,\ \BBA\ Levine}{Hartikainen
  et~al.}{2020}]{hartikainen2019dynamical}
Hartikainen, K., Geng, X., Haarnoja, T., \BBA\ Levine, S. \BBOP2020\BBCP.
\newblock \BBOQ Dynamical distance learning for semi-supervised and
  unsupervised skill discovery\BBCQ\
\newblock In {\Bem Proc. of ICLR}.

\bibitem[\protect\BCAY{Hausman, Springenberg, Wang, Heess,\ \BBA\
  Riedmiller}{Hausman et~al.}{2018}]{hausman_learning_2018}
Hausman, K., Springenberg, J.~T., Wang, Z., Heess, N., \BBA\ Riedmiller, M.~A.
  \BBOP2018\BBCP.
\newblock \BBOQ Learning an embedding space for transferable robot skills\BBCQ\
\newblock In {\Bem Proc. of ICLR}.

\bibitem[\protect\BCAY{Hermann, Hill, Green, Wang, Faulkner, Soyer, Szepesvari,
  Czarnecki, Jaderberg, Teplyashin, Wainwright, Apps, Hassabis,\ \BBA\
  Blunsom}{Hermann et~al.}{2017}]{Hermann2017}
Hermann, K.~M., Hill, F., Green, S., Wang, F., Faulkner, R., Soyer, H.,
  Szepesvari, D., Czarnecki, W.~M., Jaderberg, M., Teplyashin, D., Wainwright,
  M., Apps, C., Hassabis, D., \BBA\ Blunsom, P. \BBOP2017\BBCP.
\newblock \BBOQ {Grounded Language Learning in a Simulated 3D World}\BBCQ\
\newblock ArXiv - abs/1706.06551.

\bibitem[\protect\BCAY{Hester, Vecer{\'{\i}}k, Pietquin, Lanctot, Schaul, Piot,
  Horgan, Quan, Sendonaris, Osband, Dulac{-}Arnold, Agapiou, Leibo,\ \BBA\
  Gruslys}{Hester et~al.}{2018}]{hester2018deep}
Hester, T., Vecer{\'{\i}}k, M., Pietquin, O., Lanctot, M., Schaul, T., Piot,
  B., Horgan, D., Quan, J., Sendonaris, A., Osband, I., Dulac{-}Arnold, G.,
  Agapiou, J.~P., Leibo, J.~Z., \BBA\ Gruslys, A. \BBOP2018\BBCP.
\newblock \BBOQ Deep q-learning from demonstrations\BBCQ\
\newblock In {\Bem Proc. of AAAI}, \BPGS\ 3223--3230.

\bibitem[\protect\BCAY{Hill, Lampinen, Schneider, Clark, Botvinick,
  McClelland,\ \BBA\ Santoro}{Hill et~al.}{2020a}]{hill2019emergent}
Hill, F., Lampinen, A., Schneider, R., Clark, S., Botvinick, M., McClelland,
  J.~L., \BBA\ Santoro, A. \BBOP2020a\BBCP.
\newblock \BBOQ Emergent systematic generalization in a situated agent\BBCQ\
\newblock In {\Bem Proc. of ICLR}.

\bibitem[\protect\BCAY{Hill, Mokra, Wong,\ \BBA\ Harley}{Hill
  et~al.}{2020b}]{hill_human_2020}
Hill, F., Mokra, S., Wong, N., \BBA\ Harley, T. \BBOP2020b\BBCP.
\newblock \BBOQ Human {Instruction}-{Following} with {Deep} {Reinforcement}
  {Learning} via {Transfer}-{Learning} from {Text}\BBCQ\
\newblock ArXiv - abs/2005.09382.

\bibitem[\protect\BCAY{Hill, Tieleman, von Glehn, Wong, Merzic,\ \BBA\
  Clark}{Hill et~al.}{2021}]{hill_grounded_2020}
Hill, F., Tieleman, O., von Glehn, T., Wong, N., Merzic, H., \BBA\ Clark, S.
  \BBOP2021\BBCP.
\newblock \BBOQ Grounded language learning fast and slow\BBCQ\
\newblock In {\Bem Proc. of ICLR}.

\bibitem[\protect\BCAY{Hintze}{Hintze}{2019}]{hintze_open-endedness_2019}
Hintze, A. \BBOP2019\BBCP.
\newblock \BBOQ Open-{Endedness} for the {Sake} of {Open}-{Endedness}\BBCQ\
\newblock {\Bem Artificial Life}, {\Bem 25\/}(2), 198--206.

\bibitem[\protect\BCAY{Ho\ \BBA\ Ermon}{Ho\ \BBA\
  Ermon}{2016}]{ho2016generative}
Ho, J.\BBACOMMA\  \BBA\ Ermon, S. \BBOP2016\BBCP.
\newblock \BBOQ Generative adversarial imitation learning\BBCQ\
\newblock In {\Bem Proc. of NeurIPS}, \BPGS\ 4565--4573.

\bibitem[\protect\BCAY{Houthooft, Chen, Duan, Schulman, Turck,\ \BBA\
  Abbeel}{Houthooft et~al.}{2016}]{houthooft2016vime}
Houthooft, R., Chen, X., Duan, Y., Schulman, J., Turck, F.~D., \BBA\ Abbeel, P.
  \BBOP2016\BBCP.
\newblock \BBOQ {VIME:} variational information maximizing exploration\BBCQ\
\newblock In {\Bem Proc. of NeurIPS}, \BPGS\ 1109--1117.

\bibitem[\protect\BCAY{Jaderberg, Mnih, Czarnecki, Schaul, Leibo, Silver,\
  \BBA\ Kavukcuoglu}{Jaderberg et~al.}{2017}]{jaderberg2016reinforcement}
Jaderberg, M., Mnih, V., Czarnecki, W.~M., Schaul, T., Leibo, J.~Z., Silver,
  D., \BBA\ Kavukcuoglu, K. \BBOP2017\BBCP.
\newblock \BBOQ Reinforcement learning with unsupervised auxiliary tasks\BBCQ\
\newblock In {\Bem Proc. of ICLR}.

\bibitem[\protect\BCAY{Jiang, Gu, Murphy,\ \BBA\ Finn}{Jiang
  et~al.}{2019}]{Jiang2019}
Jiang, Y., Gu, S., Murphy, K., \BBA\ Finn, C. \BBOP2019\BBCP.
\newblock \BBOQ Language as an abstraction for hierarchical deep reinforcement
  learning\BBCQ\
\newblock In {\Bem Proc. of NeurIPS}, \BPGS\ 9414--9426.

\bibitem[\protect\BCAY{Kaelbling}{Kaelbling}{1993}]{kaelbling1993learning}
Kaelbling, L.~P. \BBOP1993\BBCP.
\newblock \BBOQ Learning to achieve goals\BBCQ\
\newblock In {\Bem IJCAI}, \BPGS\ 1094--1099. Citeseer.

\bibitem[\protect\BCAY{Kaplan\ \BBA\ Oudeyer}{Kaplan\ \BBA\
  Oudeyer}{2007}]{kaplan2007search}
Kaplan, F.\BBACOMMA\  \BBA\ Oudeyer, P.-Y. \BBOP2007\BBCP.
\newblock \BBOQ In search of the neural circuits of intrinsic motivation\BBCQ\
\newblock {\Bem Frontiers in neuroscience}, {\Bem 1}, 17.

\bibitem[\protect\BCAY{Kaplan\ \BBA\ Oudeyer}{Kaplan\ \BBA\
  Oudeyer}{2004}]{kaplan_maximizing_2004}
Kaplan, F.\BBACOMMA\  \BBA\ Oudeyer, P.-Y. \BBOP2004\BBCP.
\newblock \BBOQ Maximizing {Learning} {Progress}: {An} {Internal} {Reward}
  {System} for {Development}\BBCQ\
\newblock In {\Bem Embodied artificial intelligence}, \BPGS\ 259--270.
  Springer.

\bibitem[\protect\BCAY{Karch, Teodorescu, Hofmann, Moulin-Frier,\ \BBA\
  Oudeyer}{Karch et~al.}{2021}]{karch2021grounding}
Karch, T., Teodorescu, L., Hofmann, K., Moulin-Frier, C., \BBA\ Oudeyer, P.-Y.
  \BBOP2021\BBCP.
\newblock \BBOQ Grounding spatio-temporal language with transformers\BBCQ\
\newblock In {\Bem Proc. of NeurIPS}.

\bibitem[\protect\BCAY{Kidd\ \BBA\ Hayden}{Kidd\ \BBA\
  Hayden}{2015}]{kidd2015psychology}
Kidd, C.\BBACOMMA\  \BBA\ Hayden, B.~Y. \BBOP2015\BBCP.
\newblock \BBOQ The psychology and neuroscience of curiosity\BBCQ\
\newblock {\Bem Neuron}, {\Bem 88\/}(3), 449--460.

\bibitem[\protect\BCAY{Kim, Sano, Freitas, Haber,\ \BBA\ Yamins}{Kim
  et~al.}{2020}]{kim2020active}
Kim, K., Sano, M., Freitas, J.~D., Haber, N., \BBA\ Yamins, D. \BBOP2020\BBCP.
\newblock \BBOQ Active world model learning with progress curiosity\BBCQ\
\newblock In {\Bem Proc. of ICML}, \lowercase{\BVOL}\ 119, \BPGS\ 5306--5315.

\bibitem[\protect\BCAY{Kovač, Laversanne-Finot,\ \BBA\ Oudeyer}{Kovač
  et~al.}{2020}]{kovac2020grimgep}
Kovač, G., Laversanne-Finot, A., \BBA\ Oudeyer, P.-Y. \BBOP2020\BBCP.
\newblock \BBOQ Grimgep: Learning progress for robust goal sampling in visual
  deep reinforcement learning\BBCQ\
\newblock ArXiv - abs/2008.04388.

\bibitem[\protect\BCAY{Kulkarni, Narasimhan, Saeedi,\ \BBA\ Tenenbaum}{Kulkarni
  et~al.}{2016}]{kulkarni2016hierarchical}
Kulkarni, T.~D., Narasimhan, K., Saeedi, A., \BBA\ Tenenbaum, J.
  \BBOP2016\BBCP.
\newblock \BBOQ Hierarchical deep reinforcement learning: Integrating temporal
  abstraction and intrinsic motivation\BBCQ\
\newblock In {\Bem Proc. of NeurIPS}, \BPGS\ 3675--3683.

\bibitem[\protect\BCAY{Kumar, Kumar, Levine,\ \BBA\ Finn}{Kumar
  et~al.}{2020}]{kumar_one_2020}
Kumar, S., Kumar, A., Levine, S., \BBA\ Finn, C. \BBOP2020\BBCP.
\newblock \BBOQ One solution is not all you need: Few-shot extrapolation via
  structured maxent {RL}\BBCQ\
\newblock In {\Bem Proc. of NeurIPS}.

\bibitem[\protect\BCAY{Lanier, McAleer,\ \BBA\ Baldi}{Lanier
  et~al.}{2019}]{lanier2019curiosity}
Lanier, J.~B., McAleer, S., \BBA\ Baldi, P. \BBOP2019\BBCP.
\newblock \BBOQ Curiosity-driven multi-criteria hindsight experience
  replay\BBCQ\
\newblock ArXiv - abs/1906.03710.

\bibitem[\protect\BCAY{Lehman\ \BBA\ Stanley}{Lehman\ \BBA\
  Stanley}{2011}]{lehman2011evolving}
Lehman, J.\BBACOMMA\  \BBA\ Stanley, K.~O. \BBOP2011\BBCP.
\newblock \BBOQ Evolving a diversity of virtual creatures through novelty
  search and local competition\BBCQ\
\newblock In {\Bem Proc. of GECCO}, \BPGS\ 211--218.

\bibitem[\protect\BCAY{Levy, Platt,\ \BBA\ Saenko}{Levy
  et~al.}{2018}]{levy2018hierarchical}
Levy, A., Platt, R., \BBA\ Saenko, K. \BBOP2018\BBCP.
\newblock \BBOQ Hierarchical reinforcement learning with hindsight\BBCQ\
\newblock ArXiv - abs/1805.08180.

\bibitem[\protect\BCAY{Li, Jabri, Darrell,\ \BBA\ Agrawal}{Li
  et~al.}{2020}]{li2019towards}
Li, R., Jabri, A., Darrell, T., \BBA\ Agrawal, P. \BBOP2020\BBCP.
\newblock \BBOQ Towards practical multi-object manipulation using relational
  reinforcement learning\BBCQ\
\newblock In {\Bem 2020 IEEE International Conference on Robotics and
  Automation (ICRA)}, \BPGS\ 4051--4058. IEEE.

\bibitem[\protect\BCAY{Lindblom\ \BBA\ Ziemke}{Lindblom\ \BBA\
  Ziemke}{2003}]{lindblom_social_2003}
Lindblom, J.\BBACOMMA\  \BBA\ Ziemke, T. \BBOP2003\BBCP.
\newblock \BBOQ Social {Situatedness} of {Natural} and {Artificial}
  {Intelligence}: {Vygotsky} and {Beyond}\BBCQ\
\newblock {\Bem Adaptive Behavior}, {\Bem 11\/}(2), 79--96.

\bibitem[\protect\BCAY{Linke, Ady, White, Degris,\ \BBA\ White}{Linke
  et~al.}{2020}]{linke2019adapting}
Linke, C., Ady, N.~M., White, M., Degris, T., \BBA\ White, A. \BBOP2020\BBCP.
\newblock \BBOQ Adapting behavior via intrinsic reward: a survey and empirical
  study\BBCQ\
\newblock {\Bem Journal of Artificial Intelligence Research}, {\Bem 69},
  1287--1332.

\bibitem[\protect\BCAY{Lonini, Forestier, Teuli{\`e}re, Zhao, Shi,\ \BBA\
  Triesch}{Lonini et~al.}{2013}]{lonini2013robust}
Lonini, L., Forestier, S., Teuli{\`e}re, C., Zhao, Y., Shi, B.~E., \BBA\
  Triesch, J. \BBOP2013\BBCP.
\newblock \BBOQ Robust active binocular vision through intrinsically motivated
  learning\BBCQ\
\newblock {\Bem Frontiers in neurorobotics}, {\Bem 7}, 20.

\bibitem[\protect\BCAY{Lopes, Lang, Toussaint,\ \BBA\ Oudeyer}{Lopes
  et~al.}{2012}]{lopes2012exploration}
Lopes, M., Lang, T., Toussaint, M., \BBA\ Oudeyer, P. \BBOP2012\BBCP.
\newblock \BBOQ Exploration in model-based reinforcement learning by
  empirically estimating learning progress\BBCQ\
\newblock In {\Bem Proc. of NeurIPS}, \BPGS\ 206--214.

\bibitem[\protect\BCAY{Loynd, Fernandez, {\c{C}}elikyilmaz, Swaminathan,\ \BBA\
  Hausknecht}{Loynd et~al.}{2020}]{loynd2019working}
Loynd, R., Fernandez, R., {\c{C}}elikyilmaz, A., Swaminathan, A., \BBA\
  Hausknecht, M.~J. \BBOP2020\BBCP.
\newblock \BBOQ Working memory graphs\BBCQ\
\newblock In {\Bem Proc. of ICML}, \lowercase{\BVOL}\ 119, \BPGS\ 6404--6414.

\bibitem[\protect\BCAY{Luketina, Nardelli, Farquhar, Foerster, Andreas,
  Grefenstette, Whiteson,\ \BBA\ Rockt{\"{a}}schel}{Luketina
  et~al.}{2019}]{Luketina2019}
Luketina, J., Nardelli, N., Farquhar, G., Foerster, J.~N., Andreas, J.,
  Grefenstette, E., Whiteson, S., \BBA\ Rockt{\"{a}}schel, T. \BBOP2019\BBCP.
\newblock \BBOQ A survey of reinforcement learning informed by natural
  language\BBCQ\
\newblock In {\Bem Proc. of IJCAI}, \BPGS\ 6309--6317.

\bibitem[\protect\BCAY{Lupyan}{Lupyan}{2012}]{lupyan_what_2012}
Lupyan, G. \BBOP2012\BBCP.
\newblock \BBOQ What {Do} {Words} {Do}? {Toward} a {Theory} of
  {Language}-{Augmented} {Thought}\BBCQ\
\newblock In {\Bem Psychology of {Learning} and {Motivation}},
  \lowercase{\BVOL}~57, \BPGS\ 255--297. Elsevier.

\bibitem[\protect\BCAY{Lynch, Khansari, Xiao, Kumar, Tompson, Levine,\ \BBA\
  Sermanet}{Lynch et~al.}{2020}]{pmlr-v100-lynch20a}
Lynch, C., Khansari, M., Xiao, T., Kumar, V., Tompson, J., Levine, S., \BBA\
  Sermanet, P. \BBOP2020\BBCP.
\newblock \BBOQ Learning latent plans from play\BBCQ\
\newblock In {\Bem Proceedings of the Conference on Robot Learning},
  \lowercase{\BVOL}\ 100, \BPGS\ 1113--1132.

\bibitem[\protect\BCAY{Lynch\ \BBA\ Sermanet}{Lynch\ \BBA\
  Sermanet}{2020}]{lynch2020grounding}
Lynch, C.\BBACOMMA\  \BBA\ Sermanet, P. \BBOP2020\BBCP.
\newblock \BBOQ Grounding language in play\BBCQ\
\newblock ArXiv - abs/2005.07648.

\bibitem[\protect\BCAY{Mankowitz, {\v{Z}}{\'\i}dek, Barreto, Horgan, Hessel,
  Quan, Oh, van Hasselt, Silver,\ \BBA\ Schaul}{Mankowitz
  et~al.}{2018}]{mankowitz2018unicorn}
Mankowitz, D.~J., {\v{Z}}{\'\i}dek, A., Barreto, A., Horgan, D., Hessel, M.,
  Quan, J., Oh, J., van Hasselt, H., Silver, D., \BBA\ Schaul, T.
  \BBOP2018\BBCP.
\newblock \BBOQ Unicorn: Continual learning with a universal, off-policy
  agent\BBCQ\
\newblock ArXiv - abs/1802.08294.

\bibitem[\protect\BCAY{Martius, Der,\ \BBA\ Ay}{Martius
  et~al.}{2013}]{martius2013information}
Martius, G., Der, R., \BBA\ Ay, N. \BBOP2013\BBCP.
\newblock \BBOQ Information driven self-organization of complex robotic
  behaviors\BBCQ\
\newblock {\Bem PloS one}, {\Bem 8\/}(5), e63400.

\bibitem[\protect\BCAY{McGovern\ \BBA\ Barto}{McGovern\ \BBA\
  Barto}{2001}]{mcgovern_automatic_2001}
McGovern, A.\BBACOMMA\  \BBA\ Barto, A.~G. \BBOP2001\BBCP.
\newblock \BBOQ Automatic discovery of subgoals in reinforcement learning using
  diverse density\BBCQ\
\newblock In {\Bem Proc. of ICML}, \BPGS\ 361--368.

\bibitem[\protect\BCAY{Mirolli\ \BBA\ Parisi}{Mirolli\ \BBA\
  Parisi}{2011}]{mirolli_towards_2011}
Mirolli, M.\BBACOMMA\  \BBA\ Parisi, D. \BBOP2011\BBCP.
\newblock \BBOQ Towards a {Vygotskyan} {Cognitive} {Robotics}: {The} {Role} of
  {Language} as a {Cognitive} {Tool}\BBCQ\
\newblock {\Bem New Ideas in Psychology}, {\Bem 29\/}(3), 298--311.

\bibitem[\protect\BCAY{Mnih, Kavukcuoglu, Silver, Rusu, Veness, Bellemare,
  Graves, Riedmiller, Fidjeland, Ostrovski, et~al.}{Mnih
  et~al.}{2015}]{mnih2015human}
Mnih, V., Kavukcuoglu, K., Silver, D., Rusu, A.~A., Veness, J., Bellemare,
  M.~G., Graves, A., Riedmiller, M., Fidjeland, A.~K., Ostrovski, G., et~al.
  \BBOP2015\BBCP.
\newblock \BBOQ Human-level control through deep reinforcement learning\BBCQ\
\newblock {\Bem nature}, {\Bem 518\/}(7540), 529--533.

\bibitem[\protect\BCAY{Moerland}{Moerland}{2021}]{moerland_intersection_2021}
Moerland, T.~M. \BBOP2021\BBCP.
\newblock {\Bem The Intersection of Planning and Learning.}
\newblock Ph.D.\ thesis, Delft University of Technology, Netherlands.

\bibitem[\protect\BCAY{Mohamed\ \BBA\ Rezende}{Mohamed\ \BBA\
  Rezende}{2015}]{mohamed_variational_2015}
Mohamed, S.\BBACOMMA\  \BBA\ Rezende, D.~J. \BBOP2015\BBCP.
\newblock \BBOQ Variational information maximisation for intrinsically
  motivated reinforcement learning\BBCQ\
\newblock In {\Bem Proc. of NeurIPS}, \BPGS\ 2125--2133.

\bibitem[\protect\BCAY{Moulin-Frier, Nguyen,\ \BBA\ Oudeyer}{Moulin-Frier
  et~al.}{2014}]{moulinfriergmm}
Moulin-Frier, C., Nguyen, S.~M., \BBA\ Oudeyer, P.-Y. \BBOP2014\BBCP.
\newblock \BBOQ Self-organization of early vocal development in infants and
  machines: The role of intrinsic motivation\BBCQ\
\newblock {\Bem Frontiers in Psychology (Cognitive Science)}, {\Bem 4\/}(1006).

\bibitem[\protect\BCAY{Mouret\ \BBA\ Clune}{Mouret\ \BBA\
  Clune}{2015}]{mouret2015illuminating}
Mouret, J.-B.\BBACOMMA\  \BBA\ Clune, J. \BBOP2015\BBCP.
\newblock \BBOQ Illuminating search spaces by mapping elites\BBCQ\
\newblock ArXiv - abs/1504.04909.

\bibitem[\protect\BCAY{Nachum, Gu, Lee,\ \BBA\ Levine}{Nachum
  et~al.}{2018}]{nachum2018data}
Nachum, O., Gu, S., Lee, H., \BBA\ Levine, S. \BBOP2018\BBCP.
\newblock \BBOQ Data-efficient hierarchical reinforcement learning\BBCQ\
\newblock In {\Bem Proc. of NeurIPS}, \BPGS\ 3307--3317.

\bibitem[\protect\BCAY{Nair, Bahl, Khazatsky, Pong, Berseth,\ \BBA\
  Levine}{Nair et~al.}{2020}]{nair2020contextual}
Nair, A., Bahl, S., Khazatsky, A., Pong, V., Berseth, G., \BBA\ Levine, S.
  \BBOP2020\BBCP.
\newblock \BBOQ Contextual imagined goals for self-supervised robotic
  learning\BBCQ\
\newblock In {\Bem Conference on Robot Learning}, \BPGS\ 530--539.

\bibitem[\protect\BCAY{Nair, McGrew, Andrychowicz, Zaremba,\ \BBA\ Abbeel}{Nair
  et~al.}{2018a}]{nair2017overcoming}
Nair, A., McGrew, B., Andrychowicz, M., Zaremba, W., \BBA\ Abbeel, P.
  \BBOP2018a\BBCP.
\newblock \BBOQ Overcoming exploration in reinforcement learning with
  demonstrations\BBCQ\
\newblock In {\Bem 2018 IEEE international conference on robotics and
  automation (ICRA)}, \BPGS\ 6292--6299. IEEE.

\bibitem[\protect\BCAY{Nair, Pong, Dalal, Bahl, Lin,\ \BBA\ Levine}{Nair
  et~al.}{2018b}]{nair2018visual}
Nair, A., Pong, V., Dalal, M., Bahl, S., Lin, S., \BBA\ Levine, S.
  \BBOP2018b\BBCP.
\newblock \BBOQ Visual reinforcement learning with imagined goals\BBCQ\
\newblock In {\Bem Proc. of NeurIPS}, \BPGS\ 9209--9220.

\bibitem[\protect\BCAY{Nasiriany, Pong, Lin,\ \BBA\ Levine}{Nasiriany
  et~al.}{2019}]{nasiriany2019planning}
Nasiriany, S., Pong, V., Lin, S., \BBA\ Levine, S. \BBOP2019\BBCP.
\newblock \BBOQ Planning with goal-conditioned policies\BBCQ\
\newblock In {\Bem Proc. of NeurIPS}, \BPGS\ 14814--14825.

\bibitem[\protect\BCAY{Nguyen\ \BBA\ Oudeyer}{Nguyen\ \BBA\
  Oudeyer}{2014}]{nguyen2014socially}
Nguyen, M.\BBACOMMA\  \BBA\ Oudeyer, P.-Y. \BBOP2014\BBCP.
\newblock \BBOQ Socially guided intrinsic motivation for robot learning of
  motor skills\BBCQ\
\newblock {\Bem Autonomous Robots}, {\Bem 36\/}(3), 273--294.

\bibitem[\protect\BCAY{Nguyen-Tuong\ \BBA\ Peters}{Nguyen-Tuong\ \BBA\
  Peters}{2011}]{nguyen-tuong_model_2011}
Nguyen-Tuong, D.\BBACOMMA\  \BBA\ Peters, J. \BBOP2011\BBCP.
\newblock \BBOQ Model {Learning} for {Robot} {Control}: {A} {Survey}\BBCQ\
\newblock {\Bem Cognitive processing}, {\Bem 12\/}(4), 319--340.

\bibitem[\protect\BCAY{Oh, Singh, Lee,\ \BBA\ Kohli}{Oh
  et~al.}{2017}]{oh2017zero}
Oh, J., Singh, S.~P., Lee, H., \BBA\ Kohli, P. \BBOP2017\BBCP.
\newblock \BBOQ Zero-shot task generalization with multi-task deep
  reinforcement learning\BBCQ\
\newblock In {\Bem Proc. of ICML}, \lowercase{\BVOL}~70, \BPGS\ 2661--2670.

\bibitem[\protect\BCAY{Osa, Tangkaratt,\ \BBA\ Sugiyama}{Osa
  et~al.}{2021}]{osa_discovering_2021}
Osa, T., Tangkaratt, V., \BBA\ Sugiyama, M. \BBOP2021\BBCP.
\newblock \BBOQ Discovering {Diverse} {Solutions} in {Deep} {Reinforcement}
  {Learning}\BBCQ\
\newblock ArXiv - abs/2103.07084.

\bibitem[\protect\BCAY{Oudeyer, Kaplan,\ \BBA\ Hafner}{Oudeyer
  et~al.}{2007}]{oudeyer_intrinsic_2007}
Oudeyer, P.-Y., Kaplan, F., \BBA\ Hafner, V.~V. \BBOP2007\BBCP.
\newblock \BBOQ Intrinsic {Motivation} {Systems} for {Autonomous} {Mental}
  {Development}\BBCQ\
\newblock {\Bem IEEE transactions on evolutionary computation}, {\Bem 11\/}(2),
  265--286.

\bibitem[\protect\BCAY{Oudeyer\ \BBA\ Kaplan}{Oudeyer\ \BBA\
  Kaplan}{2007}]{oudeyer2007intrinsic}
Oudeyer, P.-Y.\BBACOMMA\  \BBA\ Kaplan, F. \BBOP2007\BBCP.
\newblock \BBOQ What is intrinsic motivation? a typology of computational
  approaches\BBCQ\
\newblock In {\Bem Frontiers in neurorobotics}, \lowercase{\BVOL}~1, \BPG~6.
  Frontiers.

\bibitem[\protect\BCAY{Oudeyer\ \BBA\ Smith}{Oudeyer\ \BBA\
  Smith}{2016}]{oudeyer2016evolution}
Oudeyer, P.-Y.\BBACOMMA\  \BBA\ Smith, L.~B. \BBOP2016\BBCP.
\newblock \BBOQ How evolution may work through curiosity-driven developmental
  process\BBCQ\
\newblock {\Bem Topics in Cognitive Science}, {\Bem 8\/}(2), 492--502.

\bibitem[\protect\BCAY{Pathak, Agrawal, Efros,\ \BBA\ Darrell}{Pathak
  et~al.}{2017}]{pathak2017curiosity}
Pathak, D., Agrawal, P., Efros, A.~A., \BBA\ Darrell, T. \BBOP2017\BBCP.
\newblock \BBOQ Curiosity-driven exploration by self-supervised
  prediction\BBCQ\
\newblock In {\Bem Proc. of ICML}, \lowercase{\BVOL}~70, \BPGS\ 2778--2787.

\bibitem[\protect\BCAY{Perez, Strub, de~Vries, Dumoulin,\ \BBA\
  Courville}{Perez et~al.}{2018}]{perez2018film}
Perez, E., Strub, F., de~Vries, H., Dumoulin, V., \BBA\ Courville, A.~C.
  \BBOP2018\BBCP.
\newblock \BBOQ Film: Visual reasoning with a general conditioning layer\BBCQ\
\newblock In {\Bem Proc. of AAAI}, \BPGS\ 3942--3951.

\bibitem[\protect\BCAY{Pitis, Chan, Zhao, Stadie,\ \BBA\ Ba}{Pitis
  et~al.}{2020}]{pitis2020maximum}
Pitis, S., Chan, H., Zhao, S., Stadie, B.~C., \BBA\ Ba, J. \BBOP2020\BBCP.
\newblock \BBOQ Maximum entropy gain exploration for long horizon multi-goal
  reinforcement learning\BBCQ\
\newblock In {\Bem Proc. of ICML}, \lowercase{\BVOL}\ 119, \BPGS\ 7750--7761.

\bibitem[\protect\BCAY{Plappert, Andrychowicz, Ray, McGrew, Baker, Powell,
  Schneider, Tobin, Chociej, Welinder, et~al.}{Plappert
  et~al.}{2018}]{plappert2018multi}
Plappert, M., Andrychowicz, M., Ray, A., McGrew, B., Baker, B., Powell, G.,
  Schneider, J., Tobin, J., Chociej, M., Welinder, P., et~al. \BBOP2018\BBCP.
\newblock \BBOQ Multi-goal reinforcement learning: Challenging robotics
  environments and request for research\BBCQ\
\newblock ArXiv - abs/1802.09464.

\bibitem[\protect\BCAY{Pong, Dalal, Lin, Nair, Bahl,\ \BBA\ Levine}{Pong
  et~al.}{2020}]{pong2019skew}
Pong, V., Dalal, M., Lin, S., Nair, A., Bahl, S., \BBA\ Levine, S.
  \BBOP2020\BBCP.
\newblock \BBOQ Skew-fit: State-covering self-supervised reinforcement
  learning\BBCQ\
\newblock In {\Bem Proc. of ICML}, \lowercase{\BVOL}\ 119, \BPGS\ 7783--7792.

\bibitem[\protect\BCAY{Portelas, Colas, Weng, Hofmann,\ \BBA\ Oudeyer}{Portelas
  et~al.}{2020a}]{portelas2020automatic}
Portelas, R., Colas, C., Weng, L., Hofmann, K., \BBA\ Oudeyer, P.
  \BBOP2020a\BBCP.
\newblock \BBOQ Automatic curriculum learning for deep {RL:} {A} short
  survey\BBCQ\
\newblock In {\Bem Proc. of IJCAI}, \BPGS\ 4819--4825.

\bibitem[\protect\BCAY{Portelas, Colas, Hofmann,\ \BBA\ Oudeyer}{Portelas
  et~al.}{2020b}]{portelas_teacher_2020}
Portelas, R., Colas, C., Hofmann, K., \BBA\ Oudeyer, P.-Y. \BBOP2020b\BBCP.
\newblock \BBOQ Teacher {Algorithms} for {Curriculum} {Learning} of {Deep} {Rl}
  in {Continuously} {Parameterized} {Environments}\BBCQ\
\newblock In {\Bem Proc. of CoRL}, \BPGS\ 835--853.

\bibitem[\protect\BCAY{Precup}{Precup}{2000a}]{precup_temporal_2000}
Precup, D. \BBOP2000a\BBCP.
\newblock {\Bem Temporal {Abstraction} in {Reinforcement} {Learning}}.
\newblock {PhD} {Thesis}, The University of Massachussetts.

\bibitem[\protect\BCAY{Precup}{Precup}{2000b}]{precup2001temporal}
Precup, D. \BBOP2000b\BBCP.
\newblock {\Bem Temporal abstraction in reinforcement learning}.
\newblock Ph.D.\ thesis, The University of Massachussetts.

\bibitem[\protect\BCAY{Racanière, Lampinen, Santoro, Reichert, Firoiu,\ \BBA\
  Lillicrap}{Racanière et~al.}{2019}]{settersolver}
Racanière, S., Lampinen, A., Santoro, A., Reichert, D., Firoiu, V., \BBA\
  Lillicrap, T. \BBOP2019\BBCP.
\newblock \BBOQ Automated curricula through setter-solver interactions\BBCQ\
\newblock ArXiv - abs/1909.12892.

\bibitem[\protect\BCAY{Raileanu\ \BBA\ Rockt{\"{a}}schel}{Raileanu\ \BBA\
  Rockt{\"{a}}schel}{2020}]{raileanu2020ride}
Raileanu, R.\BBACOMMA\  \BBA\ Rockt{\"{a}}schel, T. \BBOP2020\BBCP.
\newblock \BBOQ {RIDE:} rewarding impact-driven exploration for
  procedurally-generated environments\BBCQ\
\newblock In {\Bem Proc. of ICLR}.

\bibitem[\protect\BCAY{Ram, Leake,\ \BBA\ Leake}{Ram
  et~al.}{1995}]{ram1995goal}
Ram, A., Leake, D.~B., \BBA\ Leake, D. \BBOP1995\BBCP.
\newblock {\Bem Goal-driven learning}.
\newblock MIT press.

\bibitem[\protect\BCAY{Ramesh, Tomar,\ \BBA\ Ravindran}{Ramesh
  et~al.}{2019}]{ramesh_successor_2019}
Ramesh, R., Tomar, M., \BBA\ Ravindran, B. \BBOP2019\BBCP.
\newblock \BBOQ Successor options: An option discovery framework for
  reinforcement learning\BBCQ\
\newblock In {\Bem Proc. of IJCAI}, \BPGS\ 3304--3310.

\bibitem[\protect\BCAY{Riedmiller, Hafner, Lampe, Neunert, Degrave, de~Wiele,
  Mnih, Heess,\ \BBA\ Springenberg}{Riedmiller
  et~al.}{2018}]{riedmiller2018learning}
Riedmiller, M.~A., Hafner, R., Lampe, T., Neunert, M., Degrave, J., de~Wiele,
  T.~V., Mnih, V., Heess, N., \BBA\ Springenberg, J.~T. \BBOP2018\BBCP.
\newblock \BBOQ Learning by playing solving sparse reward tasks from
  scratch\BBCQ\
\newblock In {\Bem Proc. of ICML}, \lowercase{\BVOL}~80, \BPGS\ 4341--4350.

\bibitem[\protect\BCAY{R{\"o}der, Eppe, Nguyen,\ \BBA\ Wermter}{R{\"o}der
  et~al.}{2020}]{roder2020curious}
R{\"o}der, F., Eppe, M., Nguyen, P.~D., \BBA\ Wermter, S. \BBOP2020\BBCP.
\newblock \BBOQ Curious hierarchical actor-critic reinforcement learning\BBCQ\
\newblock In {\Bem International Conference on Artificial Neural Networks},
  \BPGS\ 408--419. Springer.

\bibitem[\protect\BCAY{Rolf\ \BBA\ Steil}{Rolf\ \BBA\
  Steil}{2013}]{rolf2013efficient}
Rolf, M.\BBACOMMA\  \BBA\ Steil, J.~J. \BBOP2013\BBCP.
\newblock \BBOQ Efficient exploratory learning of inverse kinematics on a
  bionic elephant trunk\BBCQ\
\newblock {\Bem IEEE transactions on neural networks and learning systems},
  {\Bem 25\/}(6), 1147--1160.

\bibitem[\protect\BCAY{Rolf, Steil,\ \BBA\ Gienger}{Rolf
  et~al.}{2010}]{rolf2010goal}
Rolf, M., Steil, J.~J., \BBA\ Gienger, M. \BBOP2010\BBCP.
\newblock \BBOQ Goal babbling permits direct learning of inverse
  kinematics\BBCQ\
\newblock {\Bem IEEE Transactions on Autonomous Mental Development}, {\Bem
  2\/}(3), 216--229.

\bibitem[\protect\BCAY{Ruis, Andreas, Baroni, Bouchacourt,\ \BBA\ Lake}{Ruis
  et~al.}{2020}]{ruis2020benchmark}
Ruis, L., Andreas, J., Baroni, M., Bouchacourt, D., \BBA\ Lake, B.~M.
  \BBOP2020\BBCP.
\newblock \BBOQ A benchmark for systematic generalization in grounded language
  understanding\BBCQ\
\newblock In {\Bem Proc. of NeurIPS}.

\bibitem[\protect\BCAY{Rumelhart, Smolensky, McClelland,\ \BBA\
  Hinton}{Rumelhart et~al.}{1986}]{rumelhart_sequential_1986}
Rumelhart, D.~E., Smolensky, P., McClelland, J.~L., \BBA\ Hinton, G.
  \BBOP1986\BBCP.
\newblock \BBOQ Sequential {Thought} {Processes} in {Pdp} {Models}\BBCQ\
\newblock {\Bem Parallel distributed processing: explorations in the
  microstructures of cognition}, {\Bem 2}, 3--57.

\bibitem[\protect\BCAY{Salimans, Ho, Chen, Sidor,\ \BBA\ Sutskever}{Salimans
  et~al.}{2017}]{salimans2017evolution}
Salimans, T., Ho, J., Chen, X., Sidor, S., \BBA\ Sutskever, I. \BBOP2017\BBCP.
\newblock \BBOQ Evolution strategies as a scalable alternative to reinforcement
  learning\BBCQ\
\newblock ArXiv - abs/1703.03864.

\bibitem[\protect\BCAY{Santucci, Baldassarre,\ \BBA\ Mirolli}{Santucci
  et~al.}{2016}]{santucci2016grail}
Santucci, V.~G., Baldassarre, G., \BBA\ Mirolli, M. \BBOP2016\BBCP.
\newblock \BBOQ Grail: a goal-discovering robotic architecture for
  intrinsically-motivated learning\BBCQ\
\newblock {\Bem IEEE Transactions on Cognitive and Developmental Systems},
  {\Bem 8\/}(3), 214--231.

\bibitem[\protect\BCAY{Santucci, Oudeyer, Barto,\ \BBA\ Baldassarre}{Santucci
  et~al.}{2020}]{santucci2020intrinsically}
Santucci, V.~G., Oudeyer, P.-Y., Barto, A., \BBA\ Baldassarre, G.
  \BBOP2020\BBCP.
\newblock \BBOQ Intrinsically motivated open-ended learning in autonomous
  robots\BBCQ\
\newblock In {\Bem Frontiers in Neurorobotics}, \lowercase{\BVOL}~13, \BPG\
  115. Frontiers.

\bibitem[\protect\BCAY{Schaul, Horgan, Gregor,\ \BBA\ Silver}{Schaul
  et~al.}{2015}]{schaul2015universal}
Schaul, T., Horgan, D., Gregor, K., \BBA\ Silver, D. \BBOP2015\BBCP.
\newblock \BBOQ Universal value function approximators\BBCQ\
\newblock In {\Bem Proc. of ICML}, \lowercase{\BVOL}~37, \BPGS\ 1312--1320.

\bibitem[\protect\BCAY{Schlegel, Jacobsen, Abbas, Patterson, White,\ \BBA\
  White}{Schlegel et~al.}{2021}]{schlegel_general_2021}
Schlegel, M., Jacobsen, A., Abbas, Z., Patterson, A., White, A., \BBA\ White,
  M. \BBOP2021\BBCP.
\newblock \BBOQ General value function networks\BBCQ\
\newblock {\Bem Journal of Artificial Intelligence Research}, {\Bem 70},
  497--543.

\bibitem[\protect\BCAY{Schmidhuber}{Schmidhuber}{1990}]{schmidhuber_making_1990}
Schmidhuber, J. \BBOP1990\BBCP.
\newblock \BBOQ Making the {World} {Differentiable}: {On} {Using}
  {Self}-{Supervised} {Fully} {Recurrent} {N} eu al {Networks} for {Dynamic}
  {Reinforcement} {Learning} and {Planning} in {Non}-{Stationary}
  {Environments}\BBCQ.

\bibitem[\protect\BCAY{Schmidhuber}{Schmidhuber}{1991a}]{schmidhuber1991curious}
Schmidhuber, J. \BBOP1991a\BBCP.
\newblock \BBOQ Curious model-building control systems\BBCQ\
\newblock In {\Bem Neural Networks, 1991. 1991 IEEE International Joint
  Conference on}, \BPGS\ 1458--1463. IEEE.

\bibitem[\protect\BCAY{Schmidhuber}{Schmidhuber}{1991b}]{schmidhuber1991learning}
Schmidhuber, J. \BBOP1991b\BBCP.
\newblock \BBOQ Learning to generate sub-goals for action sequences\BBCQ\
\newblock In {\Bem Artificial neural networks}, \BPGS\ 967--972.

\bibitem[\protect\BCAY{Schmidhuber}{Schmidhuber}{1991c}]{schmidhuber1991possibility}
Schmidhuber, J. \BBOP1991c\BBCP.
\newblock \BBOQ A possibility for implementing curiosity and boredom in
  model-building neural controllers\BBCQ\
\newblock In {\Bem Proc. of the international conference on simulation of
  adaptive behavior: From animals to animats}, \BPGS\ 222--227.

\bibitem[\protect\BCAY{Schrittwieser, Antonoglou, Hubert, Simonyan, Sifre,
  Schmitt, Guez, Lockhart, Hassabis, Graepel, Lillicrap,\ \BBA\
  Silver}{Schrittwieser et~al.}{2020}]{schrittwieser_mastering_2020}
Schrittwieser, J., Antonoglou, I., Hubert, T., Simonyan, K., Sifre, L.,
  Schmitt, S., Guez, A., Lockhart, E., Hassabis, D., Graepel, T., Lillicrap,
  T., \BBA\ Silver, D. \BBOP2020\BBCP.
\newblock \BBOQ Mastering {Atari}, {Go}, {Chess} and {Shogi} by {Planning} with
  a {Learned} {Model}\BBCQ\
\newblock {\Bem Nature}, {\Bem 588\/}(7839), 604--609.

\bibitem[\protect\BCAY{Sehnke, Osendorfer, R{\"u}ckstie{\ss}, Graves, Peters,\
  \BBA\ Schmidhuber}{Sehnke et~al.}{2010}]{sehnke2010parameter}
Sehnke, F., Osendorfer, C., R{\"u}ckstie{\ss}, T., Graves, A., Peters, J.,
  \BBA\ Schmidhuber, J. \BBOP2010\BBCP.
\newblock \BBOQ Parameter-exploring policy gradients\BBCQ\
\newblock {\Bem Neural Networks}, {\Bem 23\/}(4), 551--559.

\bibitem[\protect\BCAY{Sekar, Rybkin, Daniilidis, Abbeel, Hafner,\ \BBA\
  Pathak}{Sekar et~al.}{2020}]{sekar2020planning}
Sekar, R., Rybkin, O., Daniilidis, K., Abbeel, P., Hafner, D., \BBA\ Pathak, D.
  \BBOP2020\BBCP.
\newblock \BBOQ Planning to explore via self-supervised world models\BBCQ\
\newblock In {\Bem Proc. of ICML}, \lowercase{\BVOL}\ 119, \BPGS\ 8583--8592.

\bibitem[\protect\BCAY{Sharma, Gu, Levine, Kumar,\ \BBA\ Hausman}{Sharma
  et~al.}{2020}]{sharma_dynamics-aware_2020}
Sharma, A., Gu, S., Levine, S., Kumar, V., \BBA\ Hausman, K. \BBOP2020\BBCP.
\newblock \BBOQ Dynamics-aware unsupervised discovery of skills\BBCQ\
\newblock In {\Bem Proc. of ICLR}.

\bibitem[\protect\BCAY{Sigaud, Caselles-Dupr{\'e}, Colas, Akakzia, Oudeyer,\
  \BBA\ Chetouani}{Sigaud et~al.}{2021}]{sigaud2021towards}
Sigaud, O., Caselles-Dupr{\'e}, H., Colas, C., Akakzia, A., Oudeyer, P.-Y.,
  \BBA\ Chetouani, M. \BBOP2021\BBCP.
\newblock \BBOQ Towards teachable autonomous agents\BBCQ\
\newblock ArXiv - abs/2105.11977.

\bibitem[\protect\BCAY{Silver, Huang, Maddison, Guez, Sifre, Van Den~Driessche,
  Schrittwieser, Antonoglou, Panneershelvam, Lanctot, et~al.}{Silver
  et~al.}{2016}]{silver2016mastering}
Silver, D., Huang, A., Maddison, C.~J., Guez, A., Sifre, L., Van Den~Driessche,
  G., Schrittwieser, J., Antonoglou, I., Panneershelvam, V., Lanctot, M.,
  et~al. \BBOP2016\BBCP.
\newblock \BBOQ Mastering the game of go with deep neural networks and tree
  search\BBCQ\
\newblock In {\Bem nature}, \lowercase{\BVOL}\ 529, \BPGS\ 484--489. Nature
  Publishing Group.

\bibitem[\protect\BCAY{Simsek\ \BBA\ Barto}{Simsek\ \BBA\
  Barto}{2004}]{simsek_using_2004}
Simsek, {\"{O}}.\BBACOMMA\  \BBA\ Barto, A.~G. \BBOP2004\BBCP.
\newblock \BBOQ Using relative novelty to identify useful temporal abstractions
  in reinforcement learning\BBCQ\
\newblock In {\Bem Proc. of ICML}, \lowercase{\BVOL}~69.

\bibitem[\protect\BCAY{Simsek\ \BBA\ Barto}{Simsek\ \BBA\
  Barto}{2008}]{simsek_skill_2008}
Simsek, {\"{O}}.\BBACOMMA\  \BBA\ Barto, A.~G. \BBOP2008\BBCP.
\newblock \BBOQ Skill characterization based on betweenness\BBCQ\
\newblock In {\Bem Proc. of NeurIPS}, \BPGS\ 1497--1504.

\bibitem[\protect\BCAY{Singh, Lewis, Barto,\ \BBA\ Sorg}{Singh
  et~al.}{2010}]{singh2010intrinsically}
Singh, S., Lewis, R.~L., Barto, A.~G., \BBA\ Sorg, J. \BBOP2010\BBCP.
\newblock \BBOQ Intrinsically motivated reinforcement learning: An evolutionary
  perspective\BBCQ\
\newblock {\Bem IEEE Transactions on Autonomous Mental Development}, {\Bem
  2\/}(2), 70--82.

\bibitem[\protect\BCAY{Stanley}{Stanley}{2019}]{stanley_why_2019}
Stanley, K.~O. \BBOP2019\BBCP.
\newblock \BBOQ Why {Open}-{Endedness} {Matters}\BBCQ\
\newblock {\Bem Artificial life}, {\Bem 25\/}(3), 232--235.

\bibitem[\protect\BCAY{Stanley\ \BBA\ Soros}{Stanley\ \BBA\
  Soros}{2016}]{stanley_role_2016}
Stanley, K.~O.\BBACOMMA\  \BBA\ Soros, L. \BBOP2016\BBCP.
\newblock \BBOQ The {Role} of {Subjectivity} in the {Evaluation} of
  {Open}-{Endedness}\BBCQ\
\newblock In {\Bem Presentation delivered in {OEE2}: {The} {Second} {Workshop}
  on {Open}-{Ended} {Evolution}, at {ALIFE} 2016}.

\bibitem[\protect\BCAY{Stooke, Mahajan, Barros, Deck, Bauer, Sygnowski,
  Trebacz, Jaderberg, Mathieu, et~al.}{Stooke et~al.}{2021}]{team2021open}
Stooke, A., Mahajan, A., Barros, C., Deck, C., Bauer, J., Sygnowski, J.,
  Trebacz, M., Jaderberg, M., Mathieu, M., et~al. \BBOP2021\BBCP.
\newblock \BBOQ Open-ended learning leads to generally capable agents\BBCQ\
\newblock ArXiv - abs/2107.12808.

\bibitem[\protect\BCAY{Sukhbaatar, Lin, Kostrikov, Synnaeve, Szlam,\ \BBA\
  Fergus}{Sukhbaatar et~al.}{2018}]{sukhbaatar2017intrinsic}
Sukhbaatar, S., Lin, Z., Kostrikov, I., Synnaeve, G., Szlam, A., \BBA\ Fergus,
  R. \BBOP2018\BBCP.
\newblock \BBOQ Intrinsic motivation and automatic curricula via asymmetric
  self-play\BBCQ\
\newblock In {\Bem Proc. of ICLR}.

\bibitem[\protect\BCAY{Sutton\ \BBA\ Barto}{Sutton\ \BBA\
  Barto}{2018}]{sutton2018reinforcement}
Sutton, R.~S.\BBACOMMA\  \BBA\ Barto, A.~G. \BBOP2018\BBCP.
\newblock {\Bem Reinforcement learning: An introduction}.
\newblock MIT press.

\bibitem[\protect\BCAY{Sutton, Modayil, Delp, Degris, Pilarski, White,\ \BBA\
  Precup}{Sutton et~al.}{2011}]{sutton2011horde}
Sutton, R.~S., Modayil, J., Delp, M., Degris, T., Pilarski, P.~M., White, A.,
  \BBA\ Precup, D. \BBOP2011\BBCP.
\newblock \BBOQ Horde: A scalable real-time architecture for learning knowledge
  from unsupervised sensorimotor interaction\BBCQ\
\newblock In {\Bem The 10th International Conference on Autonomous Agents and
  Multiagent Systems-Volume 2}, \BPGS\ 761--768.

\bibitem[\protect\BCAY{Sutton, Precup,\ \BBA\ Singh}{Sutton
  et~al.}{1999}]{sutton_between_1999}
Sutton, R.~S., Precup, D., \BBA\ Singh, S. \BBOP1999\BBCP.
\newblock \BBOQ Between {MDPs} and {Semi}-{MDPs}: {A} {Framework} for
  {Temporal} {Abstraction} in {Reinforcement} {Learning}\BBCQ\
\newblock {\Bem Artificial intelligence}, {\Bem 112\/}(1-2), 181--211.

\bibitem[\protect\BCAY{Sutton, Precup,\ \BBA\ Singh}{Sutton
  et~al.}{1998}]{sutton1998intra}
Sutton, R.~S., Precup, D., \BBA\ Singh, S.~P. \BBOP1998\BBCP.
\newblock \BBOQ Intra-option learning about temporally abstract actions.\BBCQ\
\newblock In {\Bem ICML}, \lowercase{\BVOL}~98, \BPGS\ 556--564.

\bibitem[\protect\BCAY{Sutton\ \BBA\ Tanner}{Sutton\ \BBA\
  Tanner}{2004}]{sutton_tdnet_2004}
Sutton, R.~S.\BBACOMMA\  \BBA\ Tanner, B. \BBOP2004\BBCP.
\newblock \BBOQ Temporal-difference networks\BBCQ\
\newblock In Saul, L., Weiss, Y., \BBA\ Bottou, L.\BEDS, {\Bem Advances in
  Neural Information Processing Systems}, \lowercase{\BVOL}~17. MIT Press.

\bibitem[\protect\BCAY{Tasse, James,\ \BBA\ Rosman}{Tasse
  et~al.}{2020}]{tasse2020boolean}
Tasse, G.~N., James, S.~D., \BBA\ Rosman, B. \BBOP2020\BBCP.
\newblock \BBOQ A boolean task algebra for reinforcement learning\BBCQ\
\newblock In {\Bem Proc. of NeurIPS}.

\bibitem[\protect\BCAY{Taylor\ \BBA\ Stone}{Taylor\ \BBA\
  Stone}{2009}]{taylor2009transfer}
Taylor, M.~E.\BBACOMMA\  \BBA\ Stone, P. \BBOP2009\BBCP.
\newblock \BBOQ Transfer learning for reinforcement learning domains: A
  survey.\BBCQ\
\newblock {\Bem Journal of Machine Learning Research}, {\Bem 10\/}(7).

\bibitem[\protect\BCAY{Tomasello}{Tomasello}{1999}]{tomasello_cultural_1999}
Tomasello, M. \BBOP1999\BBCP.
\newblock {\Bem The {Cultural} {Origins} of {Human} {Cognition}}.
\newblock Harvard University Press.

\bibitem[\protect\BCAY{Tomasello}{Tomasello}{2009}]{tomasello_constructing_2009}
Tomasello, M. \BBOP2009\BBCP.
\newblock {\Bem Constructing a {Language}}.
\newblock Harvard university press.

\bibitem[\protect\BCAY{Vaswani, Shazeer, Parmar, Uszkoreit, Jones, Gomez,
  Kaiser,\ \BBA\ Polosukhin}{Vaswani et~al.}{2017}]{vaswani2017attention}
Vaswani, A., Shazeer, N., Parmar, N., Uszkoreit, J., Jones, L., Gomez, A.~N.,
  Kaiser, L., \BBA\ Polosukhin, I. \BBOP2017\BBCP.
\newblock \BBOQ Attention is all you need\BBCQ\
\newblock In {\Bem Proc. of NeurIPS}, \BPGS\ 5998--6008.

\bibitem[\protect\BCAY{Veeriah, Oh,\ \BBA\ Singh}{Veeriah
  et~al.}{2018}]{veeriah2018many}
Veeriah, V., Oh, J., \BBA\ Singh, S. \BBOP2018\BBCP.
\newblock \BBOQ Many-goals reinforcement learning\BBCQ\
\newblock ArXiv - abs/1806.09605.

\bibitem[\protect\BCAY{Venkattaramanujam, Crawford, Doan,\ \BBA\
  Precup}{Venkattaramanujam et~al.}{2019}]{venkattaramanujam2019self}
Venkattaramanujam, S., Crawford, E., Doan, T., \BBA\ Precup, D. \BBOP2019\BBCP.
\newblock \BBOQ Self-supervised learning of distance functions for
  goal-conditioned reinforcement learning\BBCQ.

\bibitem[\protect\BCAY{Vezhnevets, Osindero, Schaul, Heess, Jaderberg, Silver,\
  \BBA\ Kavukcuoglu}{Vezhnevets et~al.}{2017}]{vezhnevets2017feudal}
Vezhnevets, A.~S., Osindero, S., Schaul, T., Heess, N., Jaderberg, M., Silver,
  D., \BBA\ Kavukcuoglu, K. \BBOP2017\BBCP.
\newblock \BBOQ Feudal networks for hierarchical reinforcement learning\BBCQ\
\newblock In {\Bem Proc. of ICML}, \lowercase{\BVOL}~70, \BPGS\ 3540--3549.

\bibitem[\protect\BCAY{Vygotsky}{Vygotsky}{1934}]{vygotsky_thought_1934}
Vygotsky, L.~S. \BBOP1934\BBCP.
\newblock {\Bem Thought and {Language}}.
\newblock MIT press.

\bibitem[\protect\BCAY{Warde{-}Farley, de~Wiele, Kulkarni, Ionescu, Hansen,\
  \BBA\ Mnih}{Warde{-}Farley et~al.}{2019}]{warde2018unsupervised}
Warde{-}Farley, D., de~Wiele, T.~V., Kulkarni, T.~D., Ionescu, C., Hansen, S.,
  \BBA\ Mnih, V. \BBOP2019\BBCP.
\newblock \BBOQ Unsupervised control through non-parametric discriminative
  rewards\BBCQ\
\newblock In {\Bem Proc. of ICLR}.

\bibitem[\protect\BCAY{Whorf}{Whorf}{1956}]{whorf_language_1956}
Whorf, B.~L. \BBOP1956\BBCP.
\newblock {\Bem Language, {Thought}, and {Reality}: {Selected} {Writings} of
  {Benjamin} {Lee} {Whorf}}.
\newblock MIT press.

\bibitem[\protect\BCAY{Wierstra, Schaul, Glasmachers, Sun, Peters,\ \BBA\
  Schmidhuber}{Wierstra et~al.}{2014}]{wierstra2014natural}
Wierstra, D., Schaul, T., Glasmachers, T., Sun, Y., Peters, J., \BBA\
  Schmidhuber, J. \BBOP2014\BBCP.
\newblock \BBOQ Natural evolution strategies\BBCQ\
\newblock {\Bem The Journal of Machine Learning Research}, {\Bem 15\/}(1),
  949--980.

\bibitem[\protect\BCAY{Wood, Bruner,\ \BBA\ Ross}{Wood
  et~al.}{1976}]{wood_role_1976}
Wood, D., Bruner, J.~S., \BBA\ Ross, G. \BBOP1976\BBCP.
\newblock \BBOQ The {Role} of {Tutoring} in {Problem} {Solving}\BBCQ\
\newblock {\Bem Journal of Child Psychology and Psychiatry}, {\Bem 17\/}(2),
  89--100.

\bibitem[\protect\BCAY{Wu, Tucker,\ \BBA\ Nachum}{Wu
  et~al.}{2019}]{wu2018laplacian}
Wu, Y., Tucker, G., \BBA\ Nachum, O. \BBOP2019\BBCP.
\newblock \BBOQ The laplacian in {RL:} learning representations with efficient
  approximations\BBCQ\
\newblock In {\Bem Proc. of ICLR}.

\bibitem[\protect\BCAY{Yuan, C{\^o}t{\'e}, Fu, Lin, Pal, Bengio,\ \BBA\
  Trischler}{Yuan et~al.}{2019}]{Yuan_2019}
Yuan, X., C{\^o}t{\'e}, M.-A., Fu, J., Lin, Z., Pal, C., Bengio, Y., \BBA\
  Trischler, A. \BBOP2019\BBCP.
\newblock \BBOQ Interactive language learning by question answering\BBCQ\
\newblock In {\Bem Proc. of EMNLP}, \BPGS\ 2796--2813. Association for
  Computational Linguistics.

\bibitem[\protect\BCAY{Zhang, Abbeel,\ \BBA\ Pinto}{Zhang
  et~al.}{2020}]{zhang2020automatic}
Zhang, Y., Abbeel, P., \BBA\ Pinto, L. \BBOP2020\BBCP.
\newblock \BBOQ Automatic curriculum learning through value disagreement\BBCQ\
\newblock In {\Bem Proc. of NeurIPS}.

\bibitem[\protect\BCAY{Zhu, Mottaghi, Kolve, Lim, Gupta, Fei-Fei,\ \BBA\
  Farhadi}{Zhu et~al.}{2017}]{zhu2017target}
Zhu, Y., Mottaghi, R., Kolve, E., Lim, J.~J., Gupta, A., Fei-Fei, L., \BBA\
  Farhadi, A. \BBOP2017\BBCP.
\newblock \BBOQ Target-driven visual navigation in indoor scenes using deep
  reinforcement learning\BBCQ\
\newblock In {\Bem 2017 IEEE international conference on robotics and
  automation (ICRA)}, \BPGS\ 3357--3364. IEEE.

\bibitem[\protect\BCAY{Zlatev}{Zlatev}{2001}]{zlatev_epigenesis_2001}
Zlatev, J. \BBOP2001\BBCP.
\newblock \BBOQ The {Epigenesis} of {Meaning} in {Human} {Beings}, and
  {Possibly} in {Robots}\BBCQ\
\newblock {\Bem Minds and Machines}, {\Bem 11\/}(2), 155--195.

\end{thebibliography}

\bibliographystyle{theapa}

\end{document}